\documentclass{article}
\usepackage{amssymb}

\usepackage[final]{corl_2025} 

\usepackage[utf8]{inputenc} 
\usepackage[T1]{fontenc}    
\usepackage{hyperref}       
\usepackage{url}            
\usepackage{booktabs}       
\usepackage{amsfonts}       
\usepackage{nicefrac}       
\usepackage{microtype}      

\usepackage{graphicx}
\usepackage{amsmath}
\allowdisplaybreaks

\usepackage{amssymb}
\usepackage{mathtools}
\usepackage{amsthm}
\usepackage{enumitem}

\usepackage{wrapfig}
\usepackage[export]{adjustbox}
\usepackage{subfig}
\usepackage{subcaption} 

\usepackage{array}
\usepackage{multirow}
\usepackage{threeparttable}
\usepackage{makecell}

\usepackage{adjustbox}
\usepackage{algorithm}
\usepackage{algorithmic}
\usepackage{caption}
\usepackage{subcaption}

\usepackage{bm}
\newcommand{\btau}{\bm{\tau}}

\newcommand{\mO}{\mathcal{O}}

\newcommand{\sectionreducemargin}[1]{
\vspace{-3mm} 
\section{#1}
\vspace{-2mm} 
}

\newcommand{\subsectionreducemargin}[1]{
\vspace{-3mm} 
\subsection{#1}
\vspace{-2mm} 
}

\title{CHD: Coupled Hierarchical Diffusion for\\Long-Horizon Tasks}

\author{
  Ce Hao$^{*1}$, Anxing Xiao$^{1}$, Zhiwei Xue$^{1}$, Harold Soh$^{*1,2}$\\
  $^{1}$ School of Computing, National University of Singapore; 
  $^{2}$ Smart Systems Institute, NUS \\
  $^*$Emails: \texttt{cehao@u.nus.edu} and \texttt{harold@nus.edu.sg} \\
}

\begin{document}
\maketitle

\vspace{-8mm}

\begin{abstract}
Diffusion-based planners have shown strong performance in short-horizon tasks but often fail in complex, long-horizon settings. We trace the failure to \textit{loose coupling} between high-level (HL) sub-goal selection and low-level (LL) trajectory generation, which leads to incoherent plans and degraded performance. We propose \textbf{Coupled Hierarchical Diffusion} (CHD), a framework that models HL sub-goals and LL trajectories jointly within a unified diffusion process. A shared classifier passes LL feedback upstream so that sub-goals self-correct while sampling proceeds. This tight HL–LL coupling improves trajectory coherence and enables scalable long-horizon diffusion planning. Experiments across maze navigation, tabletop manipulation, and household environments show that CHD consistently outperforms both flat and hierarchical diffusion baselines.  \href{https://sites.google.com/view/chd2025/home}{website}
\end{abstract}
\vspace{-2mm}
\vspace{-1mm}
\sectionreducemargin{Introduction}
\label{Sec: introduction}

Diffusion models have achieved strong results in image generation~\cite{ho2020denoising}, video synthesis~\cite{ho2022video}, and protein modeling~\cite{campbell2024generative}. Recently, they’ve been applied to robot control, where diffusion-based planners generate smooth, coherent, and multi-modal trajectories~\cite{janner2022planning,ajay2022conditional, hao2025disco, zhai2025vfp}.
However, as planning horizons grow, trajectory variance increases, uncertainty compounds, and achieving high reward becomes more difficult~\cite{garrett2021integrated}. Hierarchical diffusion planners offer a promising approach towards addressing this problem by decomposing planning into high-level (HL) subgoal inference and low-level (LL) goal-conditioned trajectory generation~\cite{chen2024simple,dong2024diffuserlite}---this decomposition reduces horizon length and distribution complexity.

However, a key limitation of existing baseline hierarchical diffusion (BHD) methods is the loose coupling between the HL and LL planners. HL subgoals are generated \textit{independently} and remain \textit{fixed}, which prevents the LL planner from refining subgoals based on trajectory outcomes~\cite{li2023hierarchical,chen2024simple}. This disconnect limits coordinated planning and can lead to infeasible or sub-optimal trajectories.
This raises a central question: 
\textit{Can we couple the HL and LL planners to enable joint generation and refinement?}

\begin{figure}[t]
    \centering
    \includegraphics[width=0.85\textwidth]{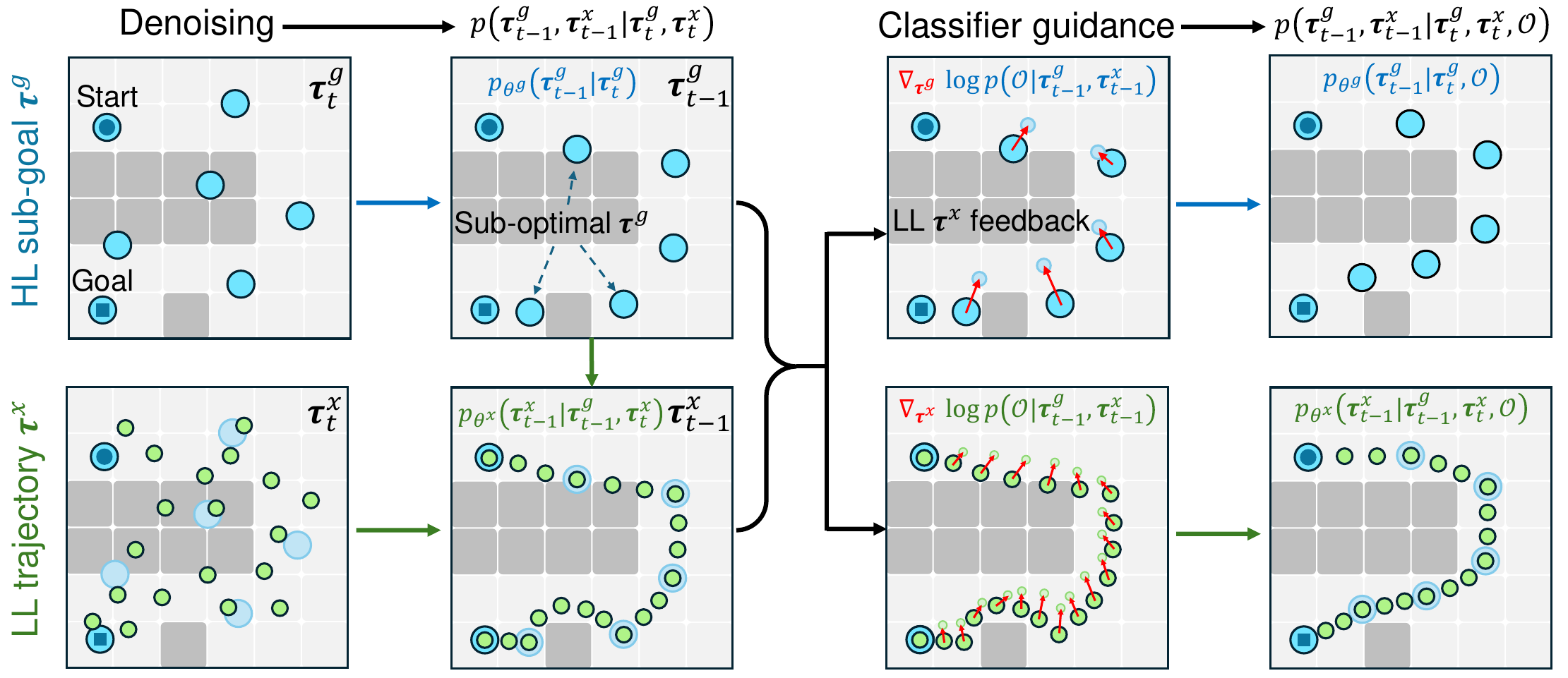}
    \caption{Illustration of our Coupled Hierarchical Diffusion (CHD). \textbf{Left}: CHD generates the joint distribution of HL and LL through the denoising process. The HL subgoals may appear reasonable, but the resulting LL trajectories are sub-optimal. \textbf{Right}: With the coupled classifier, CHD enables LL feedback to refine sub-optimal HL subgoals, leading to improved coherence and performance.}
    \label{Fig: teaser}
\vspace{-4mm}
\end{figure}

In response, we propose the \textbf{C}oupled \textbf{H}ierarchical \textbf{D}iffusion (\textbf{CHD}) algorithm for long-horizon planning (Fig.~\ref{Fig: teaser}). CHD is motivated by both practical considerations and supporting theoretical analysis (Appendix~\ref{Appendix: advantage}). 
We begin with a joint diffusion model (JDM), which jointly generates multiple variables, and adapt it into CHD by introducing three key innovations: a coupled hierarchical classifier that enables bidirectional interaction between HL subgoals and LL trajectories; an asynchronous parallel generation strategy that accelerates sampling by decoupling serial dependencies; and segment-wise generation, which reduces the planning horizon and data complexity through hierarchical decomposition.  

We evaluate CHD across challenging long-horizon tasks in both simulation and real-world settings.  
In maze navigation~\cite{fu2020d4rl}, CHD aligns HL subgoals with LL trajectory segments more effectively than both the vanilla Diffuser~\cite{janner2022planning} and the baseline SHD~\cite{li2023hierarchical}, leading to more efficient paths.  
In a robot task planning benchmark~\cite{yang2022sequence}, CHD successfully plans over 90 subtasks in a complex meal preparation scenario, handling high variance and sparse rewards, and outperforming strong baselines including Transformers~\cite{clinton2024planning}, LLMs~\cite{yang2024guiding}, and SHD.  

Finally, in real-world household settings, CHD demonstrates practical viability by generating feasible subgoals and actions for both articulated and mobile manipulation tasks. To summarize, this paper makes the following contributions:
\vspace{-1mm}
\begin{itemize}[noitemsep, leftmargin=10pt]
\item We introduce \textbf{Coupled Hierarchical Diffusion (CHD)}, a novel algorithm that jointly generates HL subgoals and LL trajectories for long-horizon planning.  
\vspace{1mm}
\item We propose a hierarchical classifier-guided diffusion process that enables LL feedback, supports asynchronous parallel generation, and reduces complexity through segment-wise planning.  
\vspace{1mm}
\item We demonstrate CHD’s effectiveness across simulated and real-world tasks, showing consistent improvements over strong baselines and highlighting the importance of HL–LL coupling.
\end{itemize}

\sectionreducemargin{Preliminaries} \label{Sec: prelim}

\vspace{2mm}
\subsectionreducemargin{Problem Formulation}
We focus on finite‑horizon, maximum‑reward trajectory optimization in a discrete‑time system~\cite{janner2022planning}.  
Let $s_k\!\in\!\mathcal{S}$ and $a_k\!\in\!\mathcal{A}$ denote the state and action at time step $k=0,\dots,H$, with dynamics 
\(
s_{k+1}=f(s_k,a_k)
\). 
During planning, we aim to generate a trajectory $\btau=(s_0,a_0,\dots,s_H,a_H)$ that maximizes the cumulative reward $\mathcal{J}(\btau) = \sum_{k=0}^{H} r(s_k,a_k)$. 

\subsectionreducemargin{Diffusion Models as Planners}
Diffusion probabilistic models~\cite{ho2020denoising} are generative models that synthesize data by iteratively denoising noisy samples from $T$ to $0$. They can also be used to generate trajectories $\btau$~\cite{janner2022planning}, where the forward process $ q(\btau_t | \btau_{t-1}) $ progressively corrupts the data (``clean'' trajectories), while the reverse process $ p_\theta(\btau_{t-1} | \btau_t) $ denoises the data. The induced marginal distribution over trajectories is 
$p_\theta(\btau_0) = \int p(\btau_T) \prod_{t=1}^T p_\theta(\btau_{t-1} | \btau_t) d\btau_{1:T},$

where $p(\btau_T) $ is a Gaussian prior and $ \btau_0 $ represents the noiseless data. The forward process $ q(\btau_t | \btau_{t-1}) $ is typically fixed and defined by Gaussian noise addition at each time-step.
Each step of the reverse process is also parameterized by a Gaussian, 
$p_\theta(\btau_{t-1} | \btau_t) = \mathcal{N}(\btau_{t-1} ; \mu_\theta(\btau_t, t), \Sigma_\theta(\btau_t, t))$, 

where the parameters $ \theta $ are optimized by minimizing a variational bound on the negative log-likelihood,
$\theta^* = \arg \min_\theta -\mathbb{E}_{\btau_0} [\log p_\theta(\btau_0)]$.

\subsectionreducemargin{Hierarchical Diffusion Planners}

In long-horizon planning, the trajectory can be naturally decomposed into a hierarchical structure~\cite{nasiriany2019planning}. Specifically, we divide the full trajectory into \( N \) segments, each with a shorter horizon \( h \), such that \( hN = H \). The high-level (HL) planner generates a sequence of subgoals \( \btau^g = \{g_i\}_{i=1}^N = (g_1, g_2, \dots, g_N) \), where \( g_i \) corresponds to the goal for the \( i \)-th segment. Let \( x = (s, a) \) denote state-action pairs. The low-level (LL) trajectory segments are defined as $\btau_{1:N}^x = \{\btau_i^x\}_{i=1}^N$ where $\btau_i^x = \{x_k\}_{k=(i-1)h}^{ih-1}$. 

We introduce a binary optimality variable \( \mathcal{O}_i = 1 \) to indicate the optimality of the \( i \)-th LL segment~\cite{levine2018reinforcement}. The hierarchical planning objective is to jointly generate HL subgoals and LL trajectories when conditioned on optimality:
\begin{equation}\label{Eqn: hier traj planning}
    p(\btau^g, \btau^x_{1:N} \mid \mathcal{O}_{1:N} = 1) 
    \propto 
    p(\btau^g)\, 
    p(\btau^x_{1:N} \mid \btau^g)\, 
    p(\mathcal{O}_{1:N} = 1 \mid \btau^x_{1:N}, \btau^g),
\end{equation}
where \( p(\btau^g) \) defines the HL subgoal distribution, \( p(\btau^x_{1:N} \mid \btau^g) \) defines the LL trajectory distribution conditioned on the subgoals, and \( p(\mathcal{O}_{1:N} = 1 \mid \btau^x_{1:N}, \btau^g) \) is a classifier that guides the joint generation.

Effective hierarchical planners for long-horizon tasks should satisfy the following key properties:
\vspace{-1mm}
\begin{enumerate}[noitemsep, leftmargin=20pt]
    \item[\textbf{P1:}] \textbf{Bi-directional Coupling.} High‑level (HL) subgoals guide low‑level (LL) trajectory generation, and LL feedback refines HL subgoals.
    \item[\textbf{P2:}] \textbf{Parallel sampling.} HL and LL levels generate subgoals and trajectories concurrently to accelerate inference.    
    \item[\textbf{P3:}] \textbf{Reduced complexity.} Hierarchical decomposition lowers the effective planning horizon and distribution complexity, improving tractability.
\end{enumerate}
\vspace{-1mm}

\begin{figure}[t]
    \centering
    \includegraphics[width=0.97\textwidth]{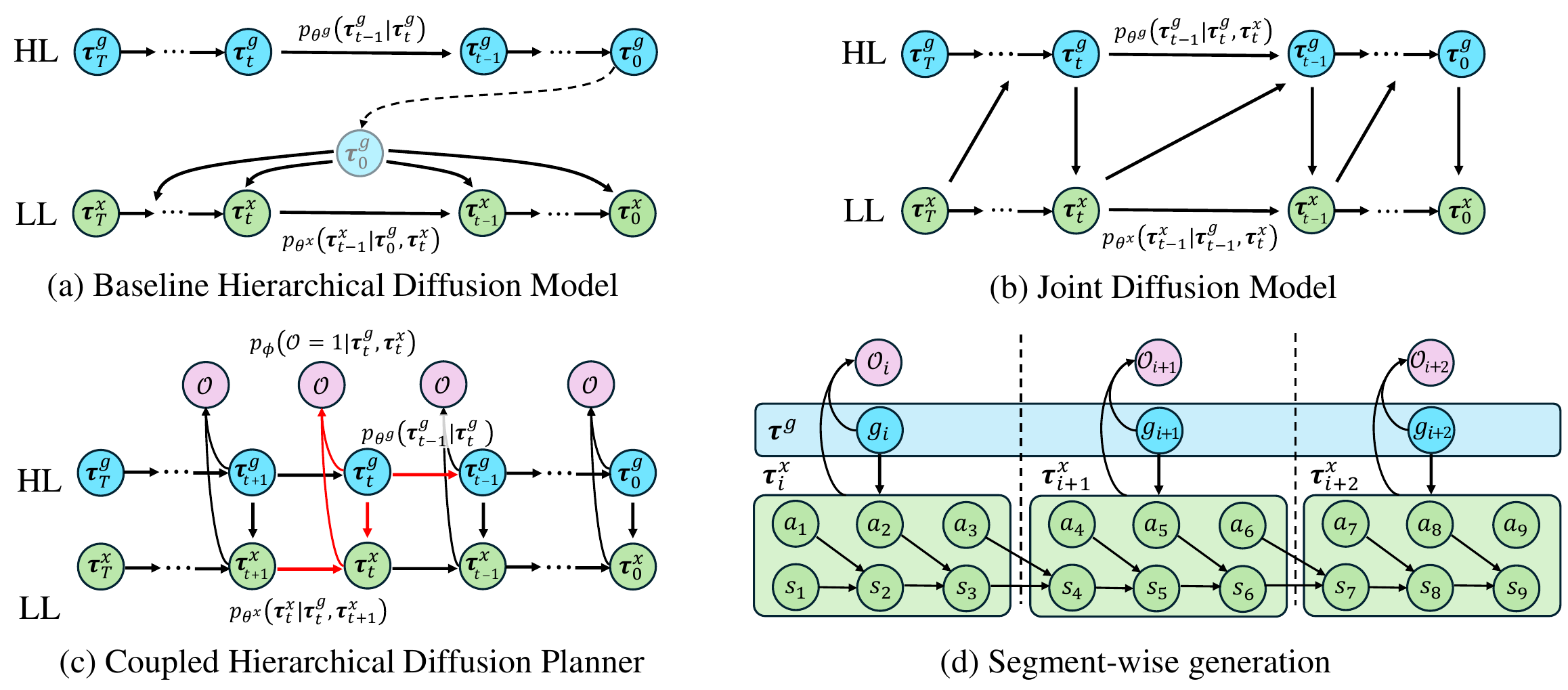}
\caption{CHD overcomes key limitations in hierarchical diffusion planning.  
(a) \textbf{BHD} plans HL subgoals and LL trajectories separately, lacking feedback and parallelism.  
(b) \textbf{JDM} enables tight HL and LL coupling but requires full joint-space diffusion.  
(c) \textbf{CHD} introduces classifier-guided LL to HL feedback and supports asynchronous, parallel generation.  
(d) Segment-wise generation further reduces horizon and complexity via localized planning.}
    \label{Fig: graph model}
\vspace{-5mm}
\end{figure}

However, existing hierarchical diffusion—what we call Baseline Hierarchical Diffuser (BHD) methods~\cite{chen2024simple, li2023hierarchical}—only satisfy P3 and fail to meet P1 and P2. As shown in Fig.~\ref{Fig: graph model}(a), during inference, BHD first generates subgoals using \(p(\btau^g)\,p(\mathcal{O}_{1:N}=1\mid\btau^g)\), where \(p(\btau^g)\) is the HL diffuser and the classifier scores the subgoal sequence. It then generates LL trajectories via the conditional diffuser \(p_{\theta^x}(\btau^x_{1:N}\mid\btau^g)\), typically implemented by subgoal inpainting. Because subgoals remain fixed before LL sampling, there is no LL to HL feedback (violating P1), and inference cannot run in parallel across levels (violating P2). Please see Appendix~\ref{Appendix: BHD} for details.

\sectionreducemargin{Method: Coupled Hierarchical Diffusion} \label{Sec: method}

In this section, we present a planning framework that satisfies the three properties outlined above. We begin from first principles, using a canonical joint diffusion model (JDM) as our foundation (Sec.~\ref{Subsec: JDM}), and then adapt it to develop the Coupled Hierarchical Diffusion (CHD) planner (Sec.~\ref{Subsec: CHD}). To realize the desired properties, CHD incorporates three core components: coupled hierarchical classifier guidance, asynchronous parallel generation, and segment-wise generation. Due to space constraints, we focus on the key ideas and refer the reader to Appendix~\ref{Appendix: JDM} and~\ref{Appendix: CHD} for  details.

Note that hierarchical diffusion involves two indices: the segment index \( i \in [1, N] \) and the diffusion step \( t \in [0, T] \). A low-level (LL) trajectory segment at step \( t \) is written as \( \btau^x_{t,i} \). For brevity, we omit the segment index and denote the full LL trajectory across segments as \( \btau^x_t = \btau^x_{t,1:N} \).

\subsectionreducemargin{Joint Diffusion Model (JDM)} \label{Subsec: JDM}

We begin by formalizing a joint diffusion model (JDM) over high-level subgoals \( \btau^g \) and low-level trajectories \( \btau^x \), treating both as part of a single generative process. In the \textbf{forward process}, the clean pair \( (\btau_0^g, \btau_0^x) \sim q(\btau^g_0, \btau^x_0) \) is corrupted independently through two separate noise processes:
\begin{align}
q(\btau^g_{1:T}, \btau^x_{1:T} \mid \btau_0^g, \btau_0^x) 
&= q(\btau^g_{1:T} \mid \btau_0^g)\, q(\btau^x_{1:T} \mid \btau_0^x), \\
q(\btau_{1:T}^{\diamond} \mid \btau_0^{\diamond}) 
&= \prod_{t=1}^T \mathcal{N}\left(\sqrt{1 - \beta_t}\, \btau_{t-1}^{\diamond},\, \beta_t \mathbf{I}^{\diamond} \right), \quad \diamond \in \{g, x\}.
\end{align}

In the \textbf{reverse process}, we model the joint generation hierarchically, as illustrated in Fig.~\ref{Fig: graph model}(b). Applying the chain rule and Markov assumptions, the reverse factorization becomes:
\begin{equation}
p(\btau^g_{0:T}, \btau^x_{0:T}) 
= p(\btau_T^g)\, p(\btau_T^x) \prod_{t=1}^T 
p_{\theta^g}(\btau_{t-1}^g \mid \btau_t^g, \btau_t^x)\,
p_{\theta^x}(\btau_{t-1}^x \mid \btau_{t-1}^g, \btau_t^x),
\end{equation}
where the initial noisy variables are drawn from Gaussian priors:  
\( p(\btau_T^g) = \mathcal{N}(\mathbf{0}, \mathbf{I}^g) \),  
\( p(\btau_T^x) = \mathcal{N}(\mathbf{0}, \mathbf{I}^x) \).
The reverse process is implemented through two coupled denoising models:  
\( p_{\theta^g} \) for HL subgoals and \( p_{\theta^x} \) for LL trajectories. Both are conditioned on one another, forming an entangled, fully joint model.

JDM is a principled and expressive formulation that naturally provides bidirectional coupling (P1) between high‑level and low‑level components. However, because it diffuses over the entire joint space, it does not reduce the planning horizon or data complexity (violating P3), nor does it allow independent, parallel sampling of HL and LL (violating P2), which limits its practical scalability.
Therefore, evaluating JDM is equivalent to the flat diffusion planner.

\subsectionreducemargin{Coupled Hierarchical Diffusion Planner} \label{Subsec: CHD}

We propose the Coupled Hierarchical Diffusion (CHD) planner, designed to satisfy all three properties (P1–P3). CHD builds on the Joint Diffusion Model (JDM) but introduces a key simplification: we remove the direct dependency of the HL reverse model on the LL trajectory. Specifically, we model the reverse process shown in Fig.~\ref{Fig: graph model}(c), where the high-level reverse step is simplified to \( p_{\theta^g}(\btau_{t-1}^g \mid \btau_t^g) \).
The full reverse process is defined as:
\begin{align}
p(\btau^g_{0:T}, \btau^x_{0:T}) &= p(\btau^g_T)\, p(\btau^x_T) \prod_{t=1}^T 
p_{\theta^g}(\btau_{t-1}^g \mid \btau_t^g)\,
p_{\theta^x}(\btau_{t-1}^x \mid \btau_{t-1}^g, \btau_t^x), \\
p_{\theta^g}(\btau_{t-1}^g \mid \btau_t^g) &= \mathcal{N}\left(\btau_{t-1}^g ;\, \mu_{\theta^g}(\btau_t^g, t),\, \Sigma_{\theta^g}(\btau_t^g, t)\right), \\
p_{\theta^x}(\btau_{t-1}^x \mid \btau_{t-1}^g, \btau_t^x) &= \mathcal{N}\left(\btau_{t-1}^x ;\, \mu_{\theta^x}(\btau_t^x, \btau_{t-1}^g, t),\, \Sigma_{\theta^x}(\btau_t^x, \btau_{t-1}^g, t)\right),
\end{align}
where \( \theta^g \) and \( \theta^x \) are the parameters of the HL and LL denoising processes, respectively.

Removing the direct dependence of the HL kernel on the LL state yields a more modular structure that we augment below with classifier guidance, asynchronous sampling, and segment‑wise planning to satisfy P1–P3. 

\textbf{Coupled hierarchical classifier guidance achieves bidirectional coupling (P1).}
Embedding CHD in the hierarchical objective of Eq.~\eqref{Eqn: hier traj planning} and applying classifier guidance yields,
\begin{equation}
\label{Eqn: chd classifier guidance}
\begin{aligned}
p&(\btau^g_{0:T}, \btau^x_{0:T} \mid \mathcal{O}_{1:N}=1) 
\propto 
p(\btau^g_{0:T}, \btau^x_{0:T})\,
p_{\phi}(\mathcal{O}_{1:N}=1 \mid \btau^g_{0:T}, \btau^x_{0:T}) \\[2pt]
&= p(\btau^g_T)\,p(\btau^x_T)
   \prod_{t=1}^{T} 
   p_{\theta^g}(\btau_{t-1}^g \mid \btau_t^g)\,
   p_{\theta^x}(\btau_{t-1}^x \mid \btau_{t-1}^g, \btau_t^x)\,
   p_{\phi}(\mathcal{O}_{1:N}=1 \mid \btau_{t-1}^g, \btau_{t-1}^x),
\end{aligned}
\end{equation}
where \(p_{\phi}(\mathcal{O}_{1:N}=1 \mid \btau_t^g, \btau_t^x)\) is the coupled hierarchical classifier with parameters \(\phi\).

As illustrated in Fig.~\ref{Fig: graph model}(c), this classifier creates a feedback channel: HL subgoals guide LL generation, while LL trajectories offer feedback through the classifier term. The latter adjusts the HL subgoals at every reverse step to improve optimality. This mutual influence realises the desired bidirectional coupling (P2) and better approximates the full-joint model (JDM) with higher optimality compared to BHD (see theoretical arguments in Appendix \ref{Appendix: advantage}).   


\textbf{Asynchronous parallel generation enables parallel sampling (P2).}  
In hierarchical planners, sampling both HL and LL levels in parallel substantially accelerates inference. However, in CHD, the probabilistic graph structure does not naturally support synchronous parallel generation because the LL reverse process depends on the HL trajectory, i.e., \( \btau^x_t \sim p_{\theta^x}(\btau^x_t \mid \btau^g_t, \btau^x_{t+1}) \). 

To enable parallelism, CHD adopts an asynchronous generation schedule by reorganizing the reverse process in Eq.~\eqref{Eqn: chd classifier guidance} as:
\begin{small}
\begin{equation} \label{Eqn: chd asy update}
\begin{aligned}
& p\left(\btau^{g}_{0:T}, \btau^{x}_{0:T} \mid \mathcal{O}_{1:N}=1\right) \propto p(\btau^g_T) p(\btau^x_T) \underbrace{p_{\theta^g}(\btau_{T-1}^g | \btau_{T}^g)}_{\mathcal{P}^g_T} \\
& \cdot \prod_{t=1}^{T-1} \bigg[ \underbrace{p_{\theta^g}(\btau_{t-1}^g | \btau_{t}^g) p_{\theta^x}(\btau_{t}^x | \btau_{t}^g, \btau_{t+1}^x)  p_{\phi}(\mathcal{O}_{1:N}=1 | \btau^g_{t}, \btau^x_{t})}_{\mathcal{P}^{g,x}_{t-1}} \bigg] 
\underbrace{p_{\theta^x}(\btau_{0}^x | \btau_{0}^g, \btau_1^x)  p_{\phi}(\mathcal{O}_{1:N}=1 | \btau^g_{0}, \btau^x_{0})}_{\mathcal{P}^x_0}
\end{aligned}
\end{equation}    
\end{small}

This decomposition separates the reverse process into three stages:
\vspace{-1mm}
\begin{enumerate}[noitemsep, leftmargin=20pt]
\item \textbf{Initialization:} (\(\mathcal{P}^g_T\)): Sample \(\btau^g_T\) and \(\btau^x_T\) from Gaussian priors, then compute the HL update \(\btau^g_{T-1}\), creating an initial stagger.
\vspace{1mm}
\item \textbf{Asynchronous core} (\(\mathcal{P}^{g,x}_{t-1}\)): For each step \(t=1,\dots,T-1\), jointly sample \((\btau^g_{t-1}, \btau^x_t)\) by updating them in parallel. The LL trajectory depends on \(\btau^x_{t+1}\) and \(\btau^g_t\), while the HL subgoal is updated independently given \(\btau^g_t\).
\vspace{1mm}
\item \textbf{Final LL step} (\(\mathcal{P}^x_0\)): Sample the last LL step \(\btau^x_0\) conditioned on \(\btau^g_0\) and \(\btau^x_1\).
\end{enumerate}
Fig.~\ref{Fig: graph model}(c) illustrates this asynchronous structure, where red arrows indicate the reverse dependencies at each step. One complication that arises is that we cannot no longer directly apply classifier guidance; the joint pair \((\btau^g_{t-1}, \btau^x_t)\) is updated at each step, but the classifier only scores \((\btau^g_t, \btau^x_t)\). Instead, we backpropagate the classifier score through the HL denoiser using the chain rule:
\begin{align}
p_{\theta^g}(\btau^g_{t-1} \mid \btau^g_t)\, p_\phi(\mathcal{O} \mid \btau^g_t, \btau^x_t) 
&\approx \mathcal{N}\left(
    \btau^g_{t-1} ;\,
    \mu_{\theta^g}(\btau^g_t, t) 
    + \lambda^g \Sigma_{\theta^g}(\btau^g_t, t)\, \mathcal{J}^{\text{Asy}}(\btau^g_t, \mu_{\theta^x}),
    \Sigma_{\theta^g}
\right) \label{Eqn: asy classifier 1} \\
\mathcal{J}^{\text{Asy}}(\btau^g_t, \mu_{\theta^x}) 
&= \nabla_{\btau^g} \log p_\phi(\mathcal{O} \mid \btau^g, \btau^x) 
\bigg|_{\substack{\btau^g = \btau^g_t \\ \btau^x = \mu_{\theta^x}}}
\cdot \frac{\partial \btau^g_t}{\partial \mu_{\theta^g}}, \label{Eqn: asy classifier 2}
\end{align}
where we have written $\mathcal{O}_{1:N}=1$ as $\mathcal{O}$, 
\(\lambda^g\) is a guidance scaling factor, and \({\partial \btau^g_t}/{\partial \mu_{\theta^g}} = \sqrt{1 - \beta_t}\) assumes fixed noise in the forward process. Through this formulation, CHD enables both HL and LL to be denoised in parallel throughout most of the reverse process, which accelerates sampling while still benefiting from classifier guidance.


\textbf{Segment-wise generation reduces planning complexity (P3).}  
In long-horizon planning, data complexity often arises from long, multi-task trajectories. CHD mitigates this by decomposing the low-level (LL) planning process into shorter, independent segments. This approach of segment-wise generation~\cite{chen2024simple} reduces the effective planning horizon and simplifies both the LL denoiser and the classifier. Specifically, we partition the LL trajectory into \( N \) segments, each of horizon \( h \), and assume conditional independence across segments. The LL reverse model and  classifier factorize as:
\begin{align}
p_{\theta^x}(\btau^x_{t-1,1:N} \mid \btau^g_{t-1}, \btau^x_{t,1:N}) & = \prod_{i=1}^{N} p_{\theta^x}(\btau^x_{t-1,i} \mid g_{t-1,i}, \btau^x_{t,i})  \label{Eqn: chd LL segment}\\
p(\mathcal{O}_{1:N}=1 \mid \btau^g_t, \btau^x_{t,1:N}) & = \prod_{i=1}^{N} p_\phi(\mathcal{O}_i=1 \mid g_{t,i}, \btau^x_{t,i}) \label{Eqn: chd classifier segment} 
\end{align}
Segment independence is an approximation, but one that significantly reduces data complexity and improves tractability in long-horizon settings.

We have focused on CHD’s core innovations above. Appendix~\ref{Appendix: CHD} details the architecture and training setup. We also explain why JDM is unsuitable for hierarchical planning (Appendix~\ref{Appendix: JDM as planner}) and provide a theoretical comparison of CHD and BHD (Appendix~\ref{Appendix: advantage}).

\sectionreducemargin{Experiments} \label{Sec: exp}

Our experiments evaluate CHD on long-horizon planning tasks, addressing three key questions:  
\textbf{(1)} Does CHD improve trajectory optimization performance?  
\textbf{(2)} Does CHD enhance hierarchical planning through LL feedback and reduced planning horizons?  
\textbf{(3)} How computationally efficient is CHD with parallel generation? We primarily test CHD on maze navigation (Sec.~\ref{Subsec: maze nav exp}) and robot task planning (Sec.~\ref{Subsec: task plan exp}), and describe a real-robot case-study involving manipulation tasks (Sec.~\ref{Subsec: real robot}). Please see Appendix~\ref{Appendix: exp} for details on the experimental setup and implementation.

\subsectionreducemargin{Maze Navigation} \label{Subsec: maze nav exp}

We begin with the Maze Navigation benchmark~\cite{fu2020d4rl}, where trajectories lie in continuous 2D space and HL subgoals correspond to key intermediate joint states between LL segments. The agent receives a reward of $+1$ upon reaching the target and $0$ otherwise; thus, the optimal policy reaches the goal in the fewest steps. This task is challenging due to sparse rewards and sub-optimal demonstration data.
We compare CHD against baseline diffusion planners, including Diffuser~\cite{janner2022planning}, Decision Diffuser (DD)~\cite{ajay2022conditional} with classifier-free guidance, and baseline hierarchical diffusion methods (BHD),  HMDI~(BHD with graph-search improved subgoals)~\cite{li2023hierarchical} and SHD~(BHD with evenly divided subgoals)~\cite{chen2024simple}. 

\begin{figure}[t]
    \centering
    \subfloat
    {
    \adjustbox{valign=m}{\includegraphics[width=0.23\textwidth]{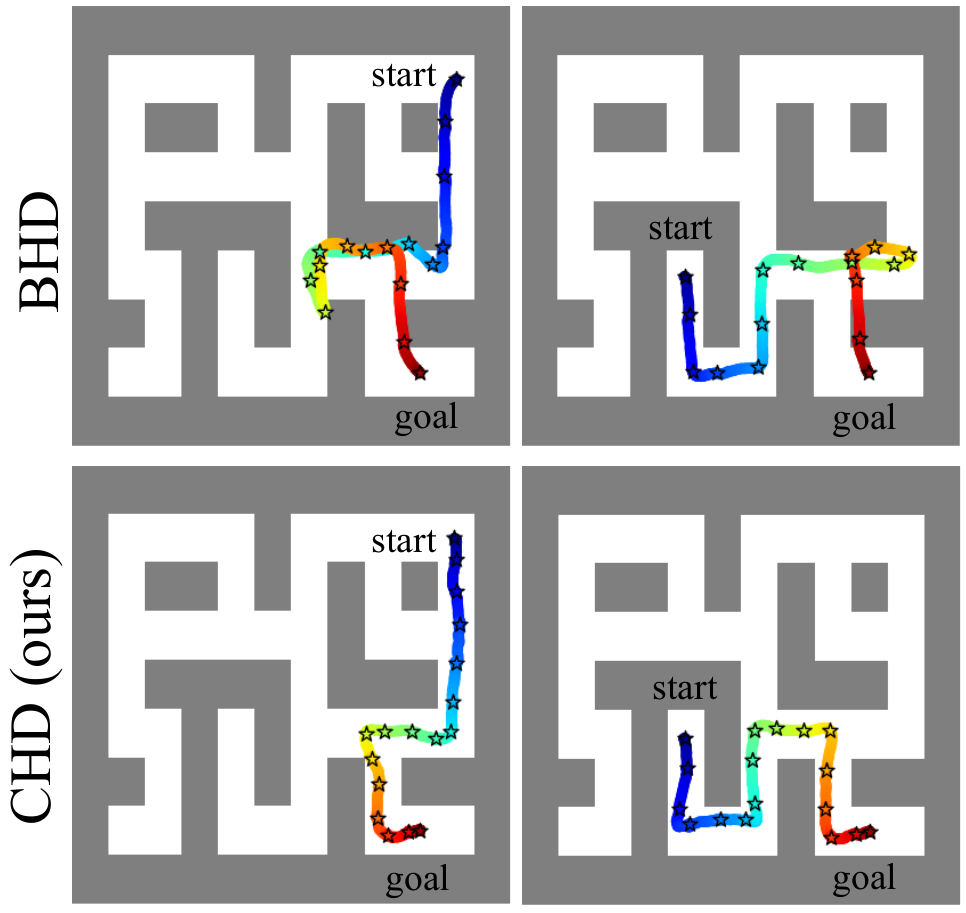}}
    }
    \hfill
    \subfloat
    {
\renewcommand\arraystretch{1.2}
\resizebox{.73\textwidth}{!}{
\begin{threeparttable}
\begin{tabular}{llcccccc}
\toprule
\multicolumn{2}{c}{\textbf{Environment}}         & \multicolumn{1}{c}{\textbf{Diffuser}} & \multicolumn{1}{c}{\textbf{DD}} & \multicolumn{1}{c}{\textbf{BHD}} & \multicolumn{1}{c}{\textbf{HMDI}} & \multicolumn{1}{c}{\textbf{SHD}} & \multicolumn{1}{c}{\textbf{CHD} (ours)} \\ \hline
Maze2D             & U-Maze            & $113.9 \pm 3.1$   & $116.2 \pm 2.7$   & $118.5 \pm 5.4$   & $120.1 \pm 2.5$   & $128.4 \pm 3.6$   & $\textbf{142.8} \pm 2.9$        \\
Maze2D             & Medium            & $121.5 \pm 2.7$   & $122.3 \pm 2.1$   & $127.1 \pm 5.3$   & $121.8 \pm 1.6$   & $135.6 \pm 3.0$   & $\textbf{149.3} \pm 3.3$        \\
Maze2D             & Large             & $123.0 \pm 6.4$   & $125.9 \pm 1.6$   & $129.0 \pm 8.5$   & $128.6 \pm 2.9$   & $155.8 \pm 2.5$   & $\textbf{179.1} \pm 4.7$        \\ \hline
\multicolumn{2}{c}{\textbf{Single-task Average}} & 119.5   & 121.5   & 124.9   & 123.5   & 139.9   & \textbf{157.1}   \\ \hline
Multi2D             & U-Maze            & $128.9 \pm 1.8$   & $128.2 \pm 2.1$   & $145.0 \pm 2.8$   & $131.3 \pm 1.8$   & $144.1 \pm 1.2$   & $\textbf{149.5} \pm 2.3$        \\
Multi2D             & Medium            & $127.2 \pm 3.4$   & $127.2 \pm 3.4$   & $130.3 \pm 4.7$   & $131.6 \pm 1.9$   & $140.2 \pm 1.6$   & $\textbf{159.9} \pm 4.1$         \\
Multi2D             & Large             & $132.1 \pm 5.8$   & $130.5 \pm 4.2$   & $150.8 \pm 6.0$   & $135.4 \pm 2.2$   & $165.5 \pm 0.6$   & $\textbf{187.4} \pm 4.8$        \\ \hline
\multicolumn{2}{c}{\textbf{Multi-task Average}}  & 129.4   & 129.5   & 142.0   & 132.8   & 149.9   & \textbf{165.6}   \\ 
\bottomrule
\end{tabular}
\end{threeparttable}
} 
    }
\caption{\textbf{Long-horizon trajectory planning in maze navigation}. \textbf{Left}: Comparision of planned trajectories, $\bigstar$ represents sub-goals. \textbf{Right}: Normalized rewards in Maze2D environments in D4RL. CHD results are calculated over 150 seeds.}
\label{Fig: maze results}
\vspace{-2mm}
\end{figure}

\textbf{Results.}  
CHD outperforms both flat and hierarchical diffusion baselines by 10–15\% in reward across different maze settings (Fig.~\ref{Fig: maze results}, right). This improvement reflects stronger imitation and trajectory optimization performance. Fig.~\ref{Fig: maze results} (left) offers a qualitative explanation for the performance gain: in BHD, subgoals are planned independently and often placed at sub-optimal corners, leading to redundant steps and collisions. In contrast, CHD couples HL and LL planning, which adapts subgoals to trajectory segments and reduces overall path length. (More details are in Appendix~\ref{Appendix: maze navigation}.)

\begin{figure}[t]
    \centering
    \includegraphics[width=0.95\linewidth]{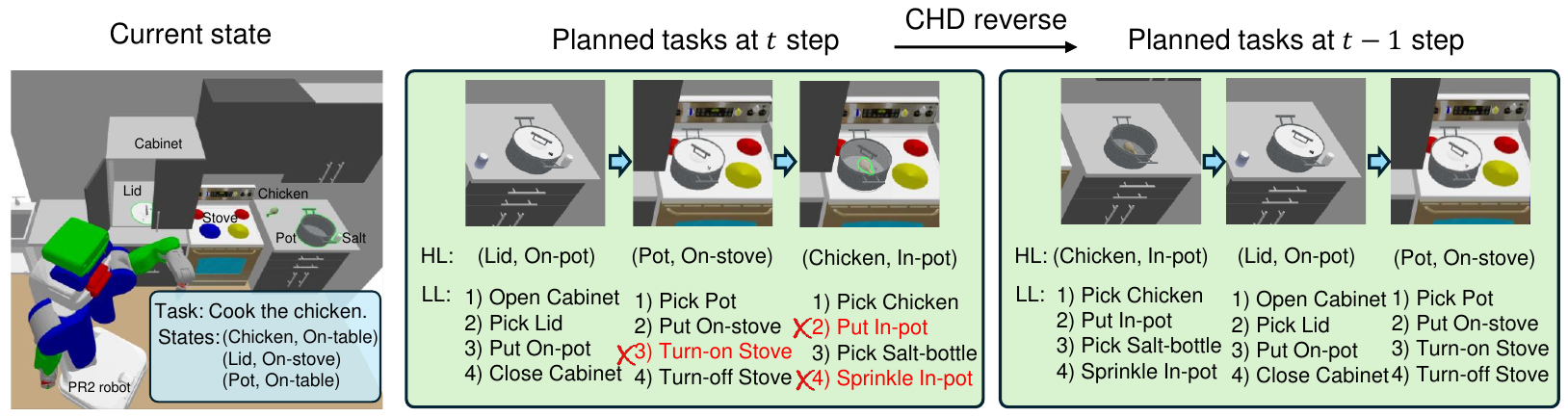}
    \caption{\textbf{Task planning experiments in Kitchen World}. Given the current state, CHD plans tasks with HL subgoal states and LL actions. During the joint reverse process, CHD can adjust the HL subgoals according to the LL actions.}
    \label{Fig: task planning}
\vspace{-4mm}
\end{figure}

\begin{figure}[t]
    \centering
    \begin{minipage}{0.49\textwidth}
        \centering
        \renewcommand\arraystretch{1.2}
        \captionof{table}{Robot Task Planning Results}
        \label{Table: task results}
        \resizebox{0.99\textwidth}{!}{ 
        \begin{threeparttable}
        \begin{tabular}{llccccc}
        \toprule
        \multicolumn{2}{l}{\textbf{Environment}} & \multicolumn{1}{c}{\textbf{VLM}} & \multicolumn{1}{c}{\textbf{Trans.}} & \multicolumn{1}{c}{\textbf{Diffuser}} & \multicolumn{1}{c}{\textbf{BHD}} & \multicolumn{1}{c}{\textbf{CHD}} 
        \\ \hline
        Single & Easy   & $4.71\pm$ \scriptsize{$0.8$} & $5.03\pm$ \scriptsize{$0.8$} & $6.34\pm$ \scriptsize{$1.4$} & $6.94\pm$ \scriptsize{$1.3$} & $\textbf{8.01}\pm$ \scriptsize{$1.6$} \\
        Single & Med. & $3.87\pm$ \scriptsize{$0.9$} & $4.37\pm$ \scriptsize{$0.7$} & $3.51\pm$ \scriptsize{$1.6$} & $5.35\pm$ \scriptsize{$1.9$} & $\textbf{5.73}\pm$ \scriptsize{$1.4$} \\ 
        Single & Hard   & $3.54\pm$ \scriptsize{$1.1$} & $4.03\pm$ \scriptsize{$0.9$} & $2.43\pm$ \scriptsize{$1.9$} & $4.17\pm$ \scriptsize{$1.7$} & $\textbf{5.10}\pm$ \scriptsize{$1.6$} \\ 
        \hline
        \multicolumn{2}{c}{\textbf{Average}}  & $3.87$      & $4.48$     & $4.09$      & $5.49$     & $\textbf{6.28}$  \\
        \hline
        Multi & Easy   & $4.45\pm$ \scriptsize{$0.8$} & $4.92\pm$ \scriptsize{$0.8$} & $5.85\pm$ \scriptsize{$1.3$} & $6.81\pm$ \scriptsize{$1.5$} & $\textbf{7.34}\pm$ \scriptsize{$1.4$} \\
        Multi & Med. & $3.89\pm$ \scriptsize{$0.8$} & $4.45\pm$ \scriptsize{$0.7$} & $3.14\pm$ \scriptsize{$1.5$} & $4.93\pm$ \scriptsize{$1.2$} & $\textbf{5.65}\pm$ \scriptsize{$1.7$} \\
        Multi & Hard   & $3.41\pm$ \scriptsize{$1.0$} & $4.02\pm$ \scriptsize{$1.0$} & $2.08\pm$ \scriptsize{$1.8$} & $3.84\pm$ \scriptsize{$1.4$} & $\textbf{4.52}\pm$ \scriptsize{$1.6$} \\ 
        \hline
        \multicolumn{2}{c}{\textbf{Average}}  & $3.91$  & $4.46$   & $3.69$    & $5.19$   & $\textbf{5.84}$  \\
        \bottomrule
        \end{tabular}
        \begin{tablenotes}
            \item[*] \small \textbf{Number of completed tasks (max $10$) $\uparrow$}. Results are averaged over $1000$ trials.
        \end{tablenotes}
        \end{threeparttable}
        }
    \end{minipage}
    \hfill
    \begin{minipage}{0.49\textwidth}
        \centering
        \includegraphics[width=0.99\textwidth]{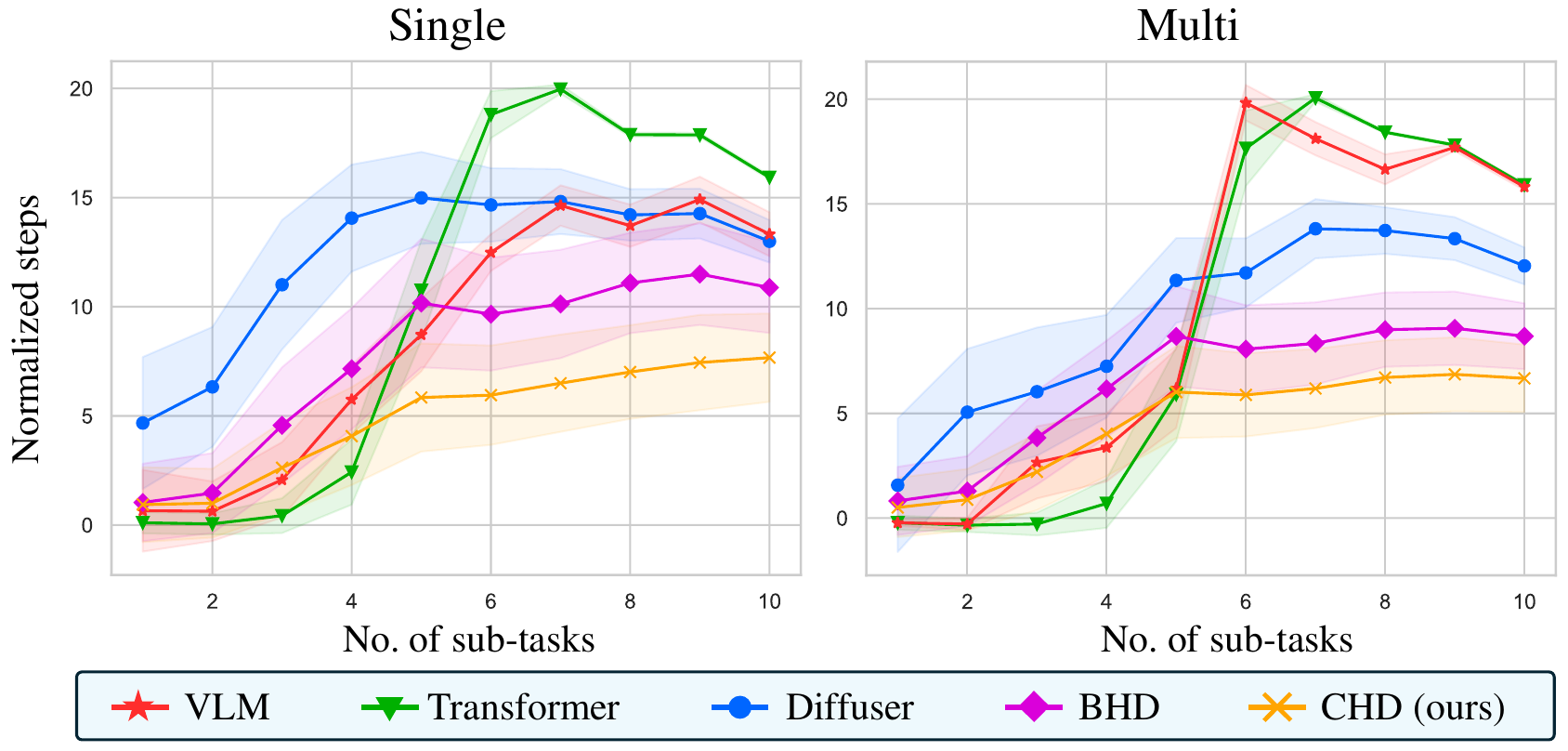}
        \captionof{figure}{\small{Normalized cumulative steps of subtasks $\downarrow$.}}
        \label{Fig: task steps} 
    \end{minipage}
\vspace{-6mm}
\end{figure}

\subsectionreducemargin{Robot Task Planning} \label{Subsec: task plan exp}

We evaluated CHD in a long-horizon robot task planning benchmark using the Kitchen-World environment~\cite{yang2024guiding}, where a mobile dual-arm robot must plan a sequence of actions to ``cook meals'' under kinematic constraints. The scene contains 20 rigid and articulated objects with randomized initial placements.
State and action spaces are discretized into semantic tokens (e.g., \texttt{(Pot, In-Cabinet)}, \texttt{(Pick, Bowl)}). We collected sub-optimal demonstrations for both single-target and multi-target tasks. These were grouped by average trajectory length into Easy (50 steps), Medium (70), and Hard (90), and further segmented into 10 sub-tasks for evaluation.

We compared CHD against several learning-based planners, including a general-purpose large language model ChatGPT-4o (VLM), Transformer, Diffuser, and BHD. For diffusion models, we applied Bits Diffusion~\cite{chen2022analog} for token generation. Performance was measured by the number of completed tasks and the normalized number of steps to complete sub-tasks.

\textbf{Results.}  
CHD achieved the highest task completion rate (Table~\ref{Table: task results}), outperforming all baselines. Unlike VLMs and Transformers, which use auto-regressive generation and often suffer from ``repetition traps'' due to context ambiguity~\cite{hiraoka2024repetition, wang2024mitigating}—diffusion-based planners iteratively refine sequences via denoising, mitigating this issue.  Diffuser outperformed VLM and Transformer, and BHD’s hierarchical structure provided further gains, particularly in medium and hard settings. However, BHD’s fixed subgoals limited recovery when early decisions became infeasible.
CHD overcame this limitation by incorporating LL feedback to refine HL subgoals during planning. Fig.~\ref{Fig: task planning} illustrates how CHD dynamically adjusts subgoals when infeasible actions are detected, improving overall robustness. 
Fig.~\ref{Fig: task steps} shows the normalized step count across sub-tasks. Transformer and VLM performed well on early sub-tasks but failed in later ones due to repetition traps. Among diffusion-based methods, CHD achieved the lowest overall step count and consistently outperformed others across both short- and long-horizon tasks. (More details are in Appendix~\ref{Appendix: task planning exp})

\subsectionreducemargin{Analysis and Discussion}

The above results show that \textbf{CHD consistently outperforms both flat and hierarchical baselines across long-horizon planning tasks}. It achieves higher success rates, better trajectory efficiency, and greater robustness in both continuous and discrete action spaces. In this section, we analyze the key design choices that contribute to these gains and examine the computational benefits of CHD’s parallel generation.

\textbf{LL feedback via coupled classifier guidance and segment-wise generation is critical for hierarchical planning.}  
We conducted three ablation studies (Table~\ref{Table: ablation results}) to assess CHD’s components. In the first two, we removed classifier guidance or used only HL-conditioned classifiers (as in BHD). Both variants saw significant drops in performance, indicating that LL feedback is important for refining HL subgoals.
In the third ablation, we replaced the segment-wise classifier (Eqn.~\eqref{Eqn: chd classifier segment}) with one conditioned on full trajectories. While this modification still outperforms BHD, it reduces performance overall due to increased distribution complexity and variance associated with conditioning on long-horizon trajectories. These results highlight the importance of both coupled guidance and segment-wise structure for effective hierarchical planning.

\textbf{CHD improves sampling efficiency through parallel generation.}  
We benchmarked training and inference times on 8$\times$NVIDIA RTX A5000 GPUs (Table~\ref{Table: test time}). CHD and BHD both train significantly faster than flat Diffuser models; the hierarhical approach reduces distribution complexity, which allowed us to reduce the number of diffusion steps from 256 to 32 while maintaining performance. 
However, at inference time, BHD requires \textit{sequential} generation of HL and LL trajectories, which increases sampling time. In contrast, CHD supports asynchronous parallel generation on GPUs by coupling HL and LL updates within the diffusion process. This leads to a 30–40\% reduction in sampling time compared to BHD, while maintaining superior planning performance.

\begin{figure}
\begin{minipage}{0.58\textwidth}
\centering
\renewcommand\arraystretch{1.2}
\captionof{table}{Comparison Against CHD Ablations}
\label{Table: ablation results}
\resizebox{.99\columnwidth}{!}{
\begin{threeparttable}
\begin{tabular}{lcccc}
\toprule
\multicolumn{1}{c}{\textbf{Environment}}         &  \multicolumn{1}{c}{\textbf{w/o Classifier}$^1$} & \multicolumn{1}{c}{\textbf{HL Classifier}$^2$} & \multicolumn{1}{c}{\textbf{w/o Seg.-wise}$^3$} & \multicolumn{1}{c}{\textbf{CHD} (ours)} \\ \hline
Maze2D      &  $122.5 \pm 6.3$  &  $138.1 \pm 3.4$  &  $142.1 \pm 4.5$   & $\textbf{157.1} \pm 3.6$        \\
Multi2D        & $142.4 \pm 4.8$  &  $146.9 \pm 5.0$    & $152.2 \pm 5.3$   & $\textbf{165.6} \pm 3.7$      \\
Single-Task        & $4.52 \pm 2.2$   & $5.53 \pm 1.7$   & $5.77 \pm 1.4$   & $\textbf{6.28} \pm 1.5$        \\
Multi-Task        & $4.16 \pm 2.4$   & $5.18 \pm 1.5$   & $5.42 \pm 1.5$   & $\textbf{5.84} \pm 1.6$        \\
\bottomrule
\end{tabular}
\begin{tablenotes}
    \small \item[$1$] Without any classifier for CHD. $^2$ Classifier only conditions on HL sub-goals, similar to BHD.
    \item[$3$] Classifier conditions on all sub-goals and whole trajectory, not segment-wise LL classifier.
    \item[$*$] Maze2D and Multi2D are average rewards in U-Maze, Medium and Large, each has 150 trials (Table~\ref{Fig: maze results}). Single-task and Multi-task are average completed tasks of Easy, Medium and Hard, each has $10,000$ trials (Table~\ref{Table: task results}).
\end{tablenotes}
\end{threeparttable}
} 
\end{minipage}
\hfill
\begin{minipage}{0.42\textwidth}
\centering
\renewcommand\arraystretch{1.2}
\captionof{table}{Computation Time Comparison}
\label{Table: test time}
\resizebox{.99\columnwidth}{!}{
\begin{threeparttable}
\begin{tabular}{lcccc}
\toprule
\multirow{2}{*}{Methods} & \multicolumn{2}{c}{Training ($h$)$^1$} & \multicolumn{2}{c}{Sampling ($s$)$^2$} \\ \cline{2-5} 
                      & Maze Nav.        & Task Plan       & Maze Nav.        & Task Plan       \\ \hline
Diffuser              & $15.0$           & $5.2$           & $1.51$            & $1.96$           \\
BHD                  & $5.0$           & $1.67$           & $0.95$            & $1.66$          \\
CHD               & $5.2$           & $1.8$           & $0.54$            & $1.13$         \\ 
\bottomrule
\end{tabular}
\begin{tablenotes}
    \item We calculate wall-clock time for training till convergence and for sampling 1 planned trajectory.
    $^1$~In the training stage, diffusion models and classifiers are trained simultaneously.
    $^2$~In the sampling stage, BHD sequentially generates HL and LL, while CHD enables parallel generation.
\end{tablenotes}
\end{threeparttable}
}
\end{minipage}
\vspace{-2mm}
\end{figure}
\newpage
\subsectionreducemargin{Real-Robot Demonstration} \label{Subsec: real robot}

\begin{figure}[t]
    \centering
    \includegraphics[width=0.99\textwidth]{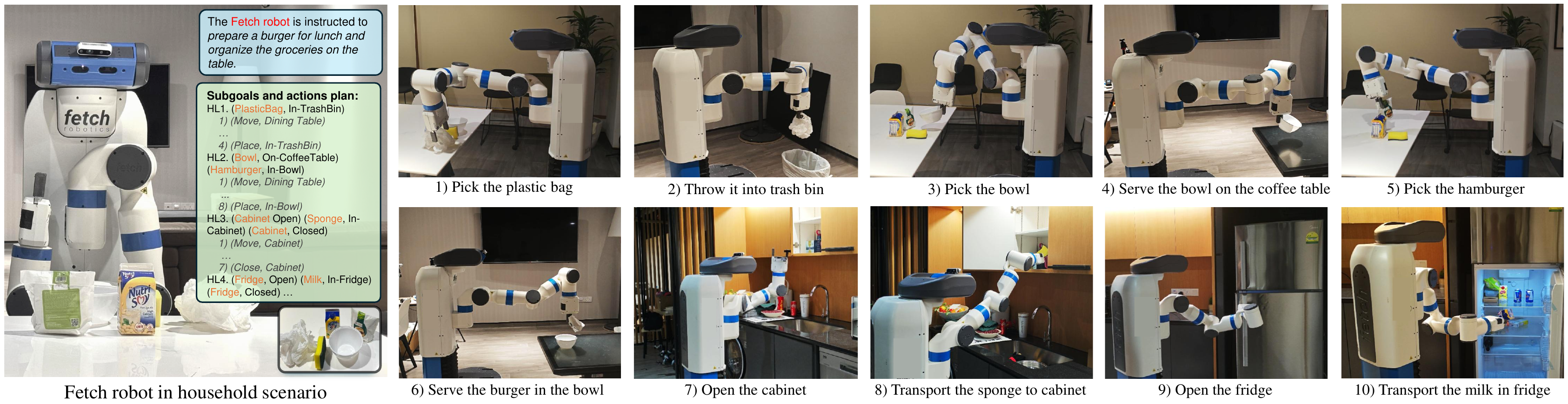}
    \caption{\textbf{Real-world task-planning demonstration.} \textbf{Left}: The Fetch mobile robot is tasked with “\textit{prepare a burger for lunch and organize the groceries on the table}.” CHD plans over 25 HL subgoals and LL actions. \textbf{Right}: Snapshots of the robot executing planned actions in a real environment. Implementation details in Appendix~\ref{Appendix: real robot}. (see supplementary \href{https://sites.google.com/view/chd2025/home}{video})}
    \label{Fig: real robot} 
\vspace{-4mm}
\end{figure}

\begin{wrapfigure}{r}{0.43\textwidth}
\vspace{-7mm}
    \centering
        \captionof{table}{\small Real-Robot Planning Success Rate $\uparrow$}
        \label{Tab: real robot success rate}
        \renewcommand\arraystretch{1.2}
        \resizebox{\linewidth}{!}{
            \begin{tabular}{lccccc}
                \toprule
                \textbf{Tasks} & \textbf{VLM} & \textbf{Trans.} & \textbf{Diffuser} & \textbf{BHD} & \textbf{CHD} \\
                \midrule
                Drop bag in bin     & $0.85$ & $\textbf{0.90}$ & $\textbf{0.90}$ & $\textbf{0.90}$ & $\textbf{0.90}$ \\
                Serve burger  & $0.35$ & $0.35$ & $0.40$ & $0.60$ & $\textbf{0.75}$ \\
                Sponge in cabinet       & $0.15$ & $0.30$ & $0.55$ & $0.65$ & $\textbf{0.80}$ \\
                Milk in fridge        & $0.20$ & $0.25$ & $0.30$ & $0.50$ & $\textbf{0.70}$ \\
                \bottomrule
            \end{tabular}
        }
\vspace{-2mm}
\end{wrapfigure}

To demonstrate CHD’s applicability to real-world long-horizon planning, we deployed it in a domestic service robot scenario (Fig.~\ref{Fig: real robot}). The robot was tasked with organizing a cluttered dining table and preparing a meal. We first collected a dataset of a single-arm robot interacting with rigid and articulated household objects, similar to Sec.~\ref{Subsec: task plan exp}, and trained CHD for task planning in this domain. Follow \cite{xiao2024robi}, we used a mobile manipulator with model-based controllers for \texttt{pick}, \texttt{place}, and \texttt{move} actions. More complex actions like \texttt{open} and \texttt{close} were trained via imitation learning~\cite{zhao2023learning} using 50 demonstration trials. 

To complete the entire task, the Fetch robot should complete four sub-tasks (Fig.~\ref{Fig: real robot}). Table~\ref{Tab: real robot success rate} presents the planning success rates of different methods.
Given that the Fetch robot is equipped with only one arm, it must adhere to hand occupancy constraints to ensure logically coherent action sequences (e.g., opening a cabinet before picking up a sponge). 
However, methods such as VLM, Transformer, and Diffuser often overlook these constraints during long-horizon planning, leading to failures, particularly when the \texttt{move} action is invoked multiple times. 
BHD addresses this limitation by predicting subgoal states, explicitly enforcing awareness of hand occupancy constraints. 
CHD jointly plans HL subgoals and LL actions through coupled diffusion, resulting in significant gains in both planning robustness and downstream policy execution.
This case study illustrates CHD’s practical applicability and its effectiveness in learning structured, constraint-aware task plans for real-world household environments.

\sectionreducemargin{Conclusion} \label{Sec: conclusion}

In this paper, we introduced Coupled Hierarchical Diffusion (CHD), a novel framework for long-horizon planning that jointly generates high-level subgoals and low-level trajectories via a coupled diffusion process. CHD addresses key limitations of existing hierarchical diffusion planners by enabling iterative feedback between planning levels, parallel sampling, and reduced complexity through segment-wise generation. 
Experiments across maze navigation and task planning demonstrate that CHD improves trajectory coherence, reward maximization, and sampling efficiency, establishing it as a strong candidate for scalable, long-horizon planning.

\clearpage

\textbf{Limitations.} While CHD performs well across long-horizon tasks, several limitations remain. First, it relies on manually defined sub-task segmentation; learning hierarchical structure automatically is an open challenge. Second, CHD assumes fixed subgoal lengths and uniform segment horizons, which may not suit tasks with varying temporal granularity. Finally, its reliance on learned policies and low-level primitives limits robustness under real-world noise and distribution shifts. Future work includes adaptive segmentation, perception integration, and safety-aware planning to improve generality and robustness.

\textbf{Acknowledgment.}
This research/project is supported by A*STAR under its National Robotics Programme (NRP) (Award M23NBK0053) and the Ministry of Education, Singapore, under the Academic Research Fund Tier 1 (T1 251RES2515).
The authors would also like to acknowledge support from Google. 



\bibliography{reference}  

\begin{thebibliography}{68}
\providecommand{\natexlab}[1]{#1}
\providecommand{\url}[1]{\texttt{#1}}
\expandafter\ifx\csname urlstyle\endcsname\relax
  \providecommand{\doi}[1]{doi: #1}\else
  \providecommand{\doi}{doi: \begingroup \urlstyle{rm}\Url}\fi

\bibitem[Ho et~al.(2020)Ho, Jain, and Abbeel]{ho2020denoising}
J.~Ho, A.~Jain, and P.~Abbeel.
\newblock Denoising diffusion probabilistic models.
\newblock \emph{Advances in neural information processing systems}, 33:\penalty0 6840--6851, 2020.

\bibitem[Ho et~al.(2022)Ho, Salimans, Gritsenko, Chan, Norouzi, and Fleet]{ho2022video}
J.~Ho, T.~Salimans, A.~Gritsenko, W.~Chan, M.~Norouzi, and D.~J. Fleet.
\newblock Video diffusion models.
\newblock \emph{Advances in Neural Information Processing Systems}, 35:\penalty0 8633--8646, 2022.

\bibitem[Campbell et~al.(2024)Campbell, Yim, Barzilay, Rainforth, and Jaakkola]{campbell2024generative}
A.~Campbell, J.~Yim, R.~Barzilay, T.~Rainforth, and T.~Jaakkola.
\newblock Generative flows on discrete state-spaces: Enabling multimodal flows with applications to protein co-design.
\newblock \emph{arXiv preprint arXiv:2402.04997}, 2024.

\bibitem[Janner et~al.(2022)Janner, Du, Tenenbaum, and Levine]{janner2022planning}
M.~Janner, Y.~Du, J.~B. Tenenbaum, and S.~Levine.
\newblock Planning with diffusion for flexible behavior synthesis.
\newblock \emph{arXiv preprint arXiv:2205.09991}, 2022.

\bibitem[Ajay et~al.(2022)Ajay, Du, Gupta, Tenenbaum, Jaakkola, and Agrawal]{ajay2022conditional}
A.~Ajay, Y.~Du, A.~Gupta, J.~Tenenbaum, T.~Jaakkola, and P.~Agrawal.
\newblock Is conditional generative modeling all you need for decision-making?
\newblock \emph{arXiv preprint arXiv:2211.15657}, 2022.

\bibitem[Hao et~al.(2025)Hao, Lin, Xue, Luo, and Soh]{hao2025disco}
C.~Hao, K.~Lin, Z.~Xue, S.~Luo, and H.~Soh.
\newblock Disco: Language-guided manipulation with diffusion policies and constrained inpainting.
\newblock \emph{IEEE Robotics and Automation Letters}, 2025.

\bibitem[Zhai and Hao(2025)]{zhai2025vfp}
X.~Zhai and C.~Hao.
\newblock Vfp: Variational flow-matching policy for multi-modal robot manipulation.
\newblock \emph{arXiv preprint arXiv:2508.01622}, 2025.

\bibitem[Garrett et~al.(2021)Garrett, Chitnis, Holladay, Kim, Silver, Kaelbling, and Lozano-P{\'e}rez]{garrett2021integrated}
C.~R. Garrett, R.~Chitnis, R.~Holladay, B.~Kim, T.~Silver, L.~P. Kaelbling, and T.~Lozano-P{\'e}rez.
\newblock Integrated task and motion planning.
\newblock \emph{Annual review of control, robotics, and autonomous systems}, 4\penalty0 (1):\penalty0 265--293, 2021.

\bibitem[Chen et~al.(2024)Chen, Deng, Kawaguchi, Gulcehre, and Ahn]{chen2024simple}
C.~Chen, F.~Deng, K.~Kawaguchi, C.~Gulcehre, and S.~Ahn.
\newblock Simple hierarchical planning with diffusion.
\newblock \emph{arXiv preprint arXiv:2401.02644}, 2024.

\bibitem[Dong et~al.(2024)Dong, Hao, Yuan, Ni, Wang, Li, and Zheng]{dong2024diffuserlite}
Z.~Dong, J.~Hao, Y.~Yuan, F.~Ni, Y.~Wang, P.~Li, and Y.~Zheng.
\newblock Diffuserlite: Towards real-time diffusion planning.
\newblock \emph{arXiv preprint arXiv:2401.15443}, 2024.

\bibitem[Li et~al.(2023)Li, Wang, Jin, and Zha]{li2023hierarchical}
W.~Li, X.~Wang, B.~Jin, and H.~Zha.
\newblock Hierarchical diffusion for offline decision making.
\newblock In \emph{International Conference on Machine Learning}, pages 20035--20064. PMLR, 2023.

\bibitem[Fu et~al.(2020)Fu, Kumar, Nachum, Tucker, and Levine]{fu2020d4rl}
J.~Fu, A.~Kumar, O.~Nachum, G.~Tucker, and S.~Levine.
\newblock D4rl: Datasets for deep data-driven reinforcement learning.
\newblock \emph{arXiv preprint arXiv:2004.07219}, 2020.

\bibitem[Yang et~al.(2022)Yang, Garrett, Lozano-P{\'e}rez, Kaelbling, and Fox]{yang2022sequence}
Z.~Yang, C.~R. Garrett, T.~Lozano-P{\'e}rez, L.~Kaelbling, and D.~Fox.
\newblock Sequence-based plan feasibility prediction for efficient task and motion planning.
\newblock \emph{arXiv preprint arXiv:2211.01576}, 2022.

\bibitem[Clinton and Lieck(2024)]{clinton2024planning}
J.~Clinton and R.~Lieck.
\newblock Planning transformer: Long-horizon offline reinforcement learning with planning tokens.
\newblock \emph{arXiv preprint arXiv:2409.09513}, 2024.

\bibitem[Yang et~al.(2024)Yang, Garrett, Fox, Lozano-P{\'e}rez, and Kaelbling]{yang2024guiding}
Z.~Yang, C.~Garrett, D.~Fox, T.~Lozano-P{\'e}rez, and L.~P. Kaelbling.
\newblock Guiding long-horizon task and motion planning with vision language models.
\newblock \emph{arXiv preprint arXiv:2410.02193}, 2024.

\bibitem[Nasiriany et~al.(2019)Nasiriany, Pong, Lin, and Levine]{nasiriany2019planning}
S.~Nasiriany, V.~Pong, S.~Lin, and S.~Levine.
\newblock Planning with goal-conditioned policies.
\newblock \emph{Advances in neural information processing systems}, 32, 2019.

\bibitem[Levine(2018)]{levine2018reinforcement}
S.~Levine.
\newblock Reinforcement learning and control as probabilistic inference: Tutorial and review.
\newblock \emph{arXiv preprint arXiv:1805.00909}, 2018.

\bibitem[Chen et~al.(2022)Chen, Zhang, and Hinton]{chen2022analog}
T.~Chen, R.~Zhang, and G.~Hinton.
\newblock Analog bits: Generating discrete data using diffusion models with self-conditioning.
\newblock \emph{arXiv preprint arXiv:2208.04202}, 2022.

\bibitem[Hiraoka and Inui(2024)]{hiraoka2024repetition}
T.~Hiraoka and K.~Inui.
\newblock Repetition neurons: How do language models produce repetitions?
\newblock \emph{arXiv preprint arXiv:2410.13497}, 2024.

\bibitem[Wang et~al.(2024)Wang, Li, Lian, Ma, Song, and Wei]{wang2024mitigating}
W.~Wang, Z.~Li, D.~Lian, C.~Ma, L.~Song, and Y.~Wei.
\newblock Mitigating the language mismatch and repetition issues in llm-based machine translation via model editing.
\newblock \emph{arXiv preprint arXiv:2410.07054}, 2024.

\bibitem[Xiao et~al.(2024)Xiao, Janaka, Hu, Gupta, Li, Yu, and Hsu]{xiao2024robi}
A.~Xiao, N.~Janaka, T.~Hu, A.~Gupta, K.~Li, C.~Yu, and D.~Hsu.
\newblock Robi butler: Remote multimodal interactions with household robot assistant.
\newblock \emph{arXiv preprint arXiv:2409.20548}, 2024.

\bibitem[Zhao et~al.(2023)Zhao, Kumar, Levine, and Finn]{zhao2023learning}
T.~Z. Zhao, V.~Kumar, S.~Levine, and C.~Finn.
\newblock Learning fine-grained bimanual manipulation with low-cost hardware.
\newblock \emph{arXiv preprint arXiv:2304.13705}, 2023.

\bibitem[Zhao et~al.(2024)Zhao, Cheng, Ding, Zhou, Zhang, Xu, and Zhao]{zhao2024survey}
Z.~Zhao, S.~Cheng, Y.~Ding, Z.~Zhou, S.~Zhang, D.~Xu, and Y.~Zhao.
\newblock A survey of optimization-based task and motion planning: From classical to learning approaches.
\newblock \emph{IEEE/ASME Transactions on Mechatronics}, 2024.

\bibitem[Urain et~al.(2024)Urain, Mandlekar, Du, Shafiullah, Xu, Fragkiadaki, Chalvatzaki, and Peters]{urain2024deep}
J.~Urain, A.~Mandlekar, Y.~Du, M.~Shafiullah, D.~Xu, K.~Fragkiadaki, G.~Chalvatzaki, and J.~Peters.
\newblock Deep generative models in robotics: A survey on learning from multimodal demonstrations.
\newblock \emph{arXiv preprint arXiv:2408.04380}, 2024.

\bibitem[Chi et~al.(2023)Chi, Xu, Feng, Cousineau, Du, Burchfiel, Tedrake, and Song]{chi2023diffusion}
C.~Chi, Z.~Xu, S.~Feng, E.~Cousineau, Y.~Du, B.~Burchfiel, R.~Tedrake, and S.~Song.
\newblock Diffusion policy: Visuomotor policy learning via action diffusion.
\newblock \emph{The International Journal of Robotics Research}, page 02783649241273668, 2023.

\bibitem[Ze et~al.(2024)Ze, Zhang, Zhang, Hu, Wang, and Xu]{ze20243d}
Y.~Ze, G.~Zhang, K.~Zhang, C.~Hu, M.~Wang, and H.~Xu.
\newblock 3d diffusion policy: Generalizable visuomotor policy learning via simple 3d representations.
\newblock In \emph{ICRA 2024 Workshop on 3D Visual Representations for Robot Manipulation}, 2024.

\bibitem[Ke et~al.(2024)Ke, Gkanatsios, and Fragkiadaki]{ke20243d}
T.-W. Ke, N.~Gkanatsios, and K.~Fragkiadaki.
\newblock 3d diffuser actor: Policy diffusion with 3d scene representations.
\newblock \emph{arXiv preprint arXiv:2402.10885}, 2024.

\bibitem[Yan et~al.(2024)Yan, Wu, and Wang]{yan2024dnact}
G.~Yan, Y.-H. Wu, and X.~Wang.
\newblock Dnact: Diffusion guided multi-task 3d policy learning.
\newblock \emph{arXiv preprint arXiv:2403.04115}, 2024.

\bibitem[Hao et~al.(2024)Hao, Lin, Luo, and Soh]{hao2024language}
C.~Hao, K.~Lin, S.~Luo, and H.~Soh.
\newblock Language-guided manipulation with diffusion policies and constrained inpainting.
\newblock \emph{arXiv preprint arXiv:2406.09767}, 2024.

\bibitem[Ha et~al.(2023)Ha, Florence, and Song]{ha2023scaling}
H.~Ha, P.~Florence, and S.~Song.
\newblock Scaling up and distilling down: Language-guided robot skill acquisition.
\newblock In \emph{Conference on Robot Learning}, pages 3766--3777. PMLR, 2023.

\bibitem[Carvalho et~al.(2023)Carvalho, Le, Baierl, Koert, and Peters]{carvalho2023motion}
J.~Carvalho, A.~T. Le, M.~Baierl, D.~Koert, and J.~Peters.
\newblock Motion planning diffusion: Learning and planning of robot motions with diffusion models.
\newblock In \emph{2023 IEEE/RSJ International Conference on Intelligent Robots and Systems (IROS)}, pages 1916--1923. IEEE, 2023.

\bibitem[Zhao et~al.(2024)Zhao, Han, Zhu, Liu, Yu, and Zhang]{zhao2024diffusion}
H.~Zhao, X.~Han, Z.~Zhu, M.~Liu, Y.~Yu, and W.~Zhang.
\newblock Diffusion-based dynamics models for long-horizon rollout in offline reinforcement learning.
\newblock \emph{arXiv preprint arXiv:2405.19189}, 2024.

\bibitem[Luo et~al.(2024)Luo, Sun, Tenenbaum, and Du]{luo2024potential}
Y.~Luo, C.~Sun, J.~B. Tenenbaum, and Y.~Du.
\newblock Potential based diffusion motion planning.
\newblock \emph{arXiv preprint arXiv:2407.06169}, 2024.

\bibitem[Xiao et~al.(2023)Xiao, Wang, Gan, and Rus]{xiao2023safediffuser}
W.~Xiao, T.-H. Wang, C.~Gan, and D.~Rus.
\newblock Safediffuser: Safe planning with diffusion probabilistic models.
\newblock \emph{arXiv preprint arXiv:2306.00148}, 2023.

\bibitem[Huang et~al.(2023)Huang, Wang, Li, Jia, Liu, Zhu, Liang, and Zhu]{huang2023diffusion}
S.~Huang, Z.~Wang, P.~Li, B.~Jia, T.~Liu, Y.~Zhu, W.~Liang, and S.-C. Zhu.
\newblock Diffusion-based generation, optimization, and planning in 3d scenes.
\newblock In \emph{Proceedings of the IEEE/CVF Conference on Computer Vision and Pattern Recognition}, pages 16750--16761, 2023.

\bibitem[He et~al.(2023)He, Bai, Xu, Yang, Zhang, Wang, Zhao, and Li]{he2023diffusion}
H.~He, C.~Bai, K.~Xu, Z.~Yang, W.~Zhang, D.~Wang, B.~Zhao, and X.~Li.
\newblock Diffusion model is an effective planner and data synthesizer for multi-task reinforcement learning.
\newblock \emph{Advances in neural information processing systems}, 36:\penalty0 64896--64917, 2023.

\bibitem[Yang et~al.(2023)Yang, Xu, Wu, Gao, Chang, and Gao]{yang2023planning}
C.-F. Yang, H.~Xu, T.-L. Wu, X.~Gao, K.-W. Chang, and F.~Gao.
\newblock Planning as in-painting: A diffusion-based embodied task planning framework for environments under uncertainty.
\newblock \emph{arXiv preprint arXiv:2312.01097}, 2023.

\bibitem[Nisonoff et~al.(2024)Nisonoff, Xiong, Allenspach, and Listgarten]{nisonoff2024unlocking}
H.~Nisonoff, J.~Xiong, S.~Allenspach, and J.~Listgarten.
\newblock Unlocking guidance for discrete state-space diffusion and flow models.
\newblock \emph{arXiv preprint arXiv:2406.01572}, 2024.

\bibitem[Zhu et~al.(2023)Zhu, Zhao, He, Zhong, Zhang, Guo, Chen, and Zhang]{zhu2023diffusion}
Z.~Zhu, H.~Zhao, H.~He, Y.~Zhong, S.~Zhang, H.~Guo, T.~Chen, and W.~Zhang.
\newblock Diffusion models for reinforcement learning: A survey.
\newblock \emph{arXiv preprint arXiv:2311.01223}, 2023.

\bibitem[Zhang et~al.(2024)Zhang, Guan, Zhao, Li, Li, Zeng, Sun, Chen, Wei, Li, et~al.]{zhang2024preferred}
T.~Zhang, J.~Guan, L.~Zhao, Y.~Li, D.~Li, Z.~Zeng, L.~Sun, Y.~Chen, X.~Wei, L.~Li, et~al.
\newblock Preferred-action-optimized diffusion policies for offline reinforcement learning.
\newblock \emph{arXiv preprint arXiv:2405.18729}, 2024.

\bibitem[Wang et~al.(2022)Wang, Hunt, and Zhou]{wang2022diffusion}
Z.~Wang, J.~J. Hunt, and M.~Zhou.
\newblock Diffusion policies as an expressive policy class for offline reinforcement learning.
\newblock \emph{arXiv preprint arXiv:2208.06193}, 2022.

\bibitem[Ren et~al.(2024)Ren, Lidard, Ankile, Simeonov, Agrawal, Majumdar, Burchfiel, Dai, and Simchowitz]{ren2024diffusion}
A.~Z. Ren, J.~Lidard, L.~L. Ankile, A.~Simeonov, P.~Agrawal, A.~Majumdar, B.~Burchfiel, H.~Dai, and M.~Simchowitz.
\newblock Diffusion policy policy optimization.
\newblock \emph{arXiv preprint arXiv:2409.00588}, 2024.

\bibitem[Wang et~al.(2024)Wang, Wang, Jiang, Zou, Liu, Song, Wang, Xiao, Wu, Duan, et~al.]{wang2024diffusion}
Y.~Wang, L.~Wang, Y.~Jiang, W.~Zou, T.~Liu, X.~Song, W.~Wang, L.~Xiao, J.~Wu, J.~Duan, et~al.
\newblock Diffusion actor-critic with entropy regulator.
\newblock \emph{arXiv preprint arXiv:2405.15177}, 2024.

\bibitem[Psenka et~al.(2023)Psenka, Escontrela, Abbeel, and Ma]{psenka2023learning}
M.~Psenka, A.~Escontrela, P.~Abbeel, and Y.~Ma.
\newblock Learning a diffusion model policy from rewards via q-score matching.
\newblock \emph{arXiv preprint arXiv:2312.11752}, 2023.

\bibitem[Ada et~al.(2024)Ada, Oztop, and Ugur]{ada2024diffusion}
S.~E. Ada, E.~Oztop, and E.~Ugur.
\newblock Diffusion policies for out-of-distribution generalization in offline reinforcement learning.
\newblock \emph{IEEE Robotics and Automation Letters}, 2024.

\bibitem[Hansen-Estruch et~al.(2023)Hansen-Estruch, Kostrikov, Janner, Kuba, and Levine]{hansen2023idql}
P.~Hansen-Estruch, I.~Kostrikov, M.~Janner, J.~G. Kuba, and S.~Levine.
\newblock Idql: Implicit q-learning as an actor-critic method with diffusion policies.
\newblock \emph{arXiv preprint arXiv:2304.10573}, 2023.

\bibitem[Zhang et~al.(2024)Zhang, Cheng, Cao, and Wang]{zhang2024offline}
J.~Zhang, Y.~Cheng, S.~Cao, and X.~Wang.
\newblock Offline reinforcement learning with reverse diffusion guide policy.
\newblock \emph{IEEE Transactions on Industrial Informatics}, 2024.

\bibitem[Pateria et~al.(2021)Pateria, Subagdja, Tan, and Quek]{pateria2021hierarchical}
S.~Pateria, B.~Subagdja, A.-h. Tan, and C.~Quek.
\newblock Hierarchical reinforcement learning: A comprehensive survey.
\newblock \emph{ACM Computing Surveys (CSUR)}, 54\penalty0 (5):\penalty0 1--35, 2021.

\bibitem[Chen et~al.(2023)Chen, Xiao, and Hsu]{chen2023llm}
S.~Chen, A.~Xiao, and D.~Hsu.
\newblock Llm-state: Expandable state representation for long-horizon task planning in the open world.
\newblock \emph{arXiv preprint arXiv:2311.17406}, 2023.

\bibitem[Aceituno and Rodriguez(2022)]{aceituno2022hierarchical}
B.~Aceituno and A.~Rodriguez.
\newblock A hierarchical framework for long horizon planning of object-contact trajectories.
\newblock In \emph{2022 IEEE/RSJ International Conference on Intelligent Robots and Systems (IROS)}, pages 189--196. IEEE, 2022.

\bibitem[Pertsch et~al.(2020)Pertsch, Rybkin, Ebert, Zhou, Jayaraman, Finn, and Levine]{pertsch2020long}
K.~Pertsch, O.~Rybkin, F.~Ebert, S.~Zhou, D.~Jayaraman, C.~Finn, and S.~Levine.
\newblock Long-horizon visual planning with goal-conditioned hierarchical predictors.
\newblock \emph{Advances in Neural Information Processing Systems}, 33:\penalty0 17321--17333, 2020.

\bibitem[Feng et~al.(2024)Feng, Luan, Ma, and Soh]{feng2024diffusion}
Z.~Feng, H.~Luan, K.~Y. Ma, and H.~Soh.
\newblock Diffusion meets options: Hierarchical generative skill composition for temporally-extended tasks.
\newblock \emph{arXiv preprint arXiv:2410.02389}, 2024.

\bibitem[Huang et~al.(2024)Huang, Wang, Li, Zhang, and Fei-Fei]{huang2024rekep}
W.~Huang, C.~Wang, Y.~Li, R.~Zhang, and L.~Fei-Fei.
\newblock Rekep: Spatio-temporal reasoning of relational keypoint constraints for robotic manipulation.
\newblock \emph{arXiv preprint arXiv:2409.01652}, 2024.

\bibitem[Paulius et~al.(2023)Paulius, Agostini, and Lee]{paulius2023long}
D.~Paulius, A.~Agostini, and D.~Lee.
\newblock Long-horizon planning and execution with functional object-oriented networks.
\newblock \emph{IEEE Robotics and Automation Letters}, 8\penalty0 (8):\penalty0 4513--4520, 2023.

\bibitem[Huang et~al.(2024)Huang, Lin, Yang, and Berenson]{huang2024subgoal}
Z.~Huang, Y.~Lin, F.~Yang, and D.~Berenson.
\newblock Subgoal diffuser: Coarse-to-fine subgoal generation to guide model predictive control for robot manipulation.
\newblock \emph{arXiv preprint arXiv:2403.13085}, 2024.

\bibitem[Liang et~al.(2024)Liang, Mu, Ma, Tomizuka, Ding, and Luo]{liang2024skilldiffuser}
Z.~Liang, Y.~Mu, H.~Ma, M.~Tomizuka, M.~Ding, and P.~Luo.
\newblock Skilldiffuser: Interpretable hierarchical planning via skill abstractions in diffusion-based task execution.
\newblock In \emph{Proceedings of the IEEE/CVF Conference on Computer Vision and Pattern Recognition}, pages 16467--16476, 2024.

\bibitem[Kim et~al.(2024{\natexlab{a}})Kim, Yoo, and Woo]{kim2024robust}
W.~K. Kim, M.~Yoo, and H.~Woo.
\newblock Robust policy learning via offline skill diffusion.
\newblock In \emph{Proceedings of the AAAI Conference on Artificial Intelligence}, volume~38, pages 13177--13184, 2024{\natexlab{a}}.

\bibitem[Kim et~al.(2024{\natexlab{b}})Kim, Choi, Matsunaga, and Kim]{kim2024stitching}
S.~Kim, Y.~Choi, D.~E. Matsunaga, and K.-E. Kim.
\newblock Stitching sub-trajectories with conditional diffusion model for goal-conditioned offline rl.
\newblock In \emph{Proceedings of the AAAI Conference on Artificial Intelligence}, volume~38, pages 13160--13167, 2024{\natexlab{b}}.

\bibitem[Wu et~al.(2024)Wu, Ye, Natarajan, and Gombolay]{wu2024diffusion}
Z.~Wu, S.~Ye, M.~Natarajan, and M.~C. Gombolay.
\newblock Diffusion-reinforcement learning hierarchical motion planning in adversarial multi-agent games.
\newblock \emph{arXiv preprint arXiv:2403.10794}, 2024.

\bibitem[Zhang et~al.(2023)Zhang, Jiang, Jiang, and Jiang]{zhang2023hierarchical}
C.~Zhang, D.~Jiang, K.~Jiang, and B.~Jiang.
\newblock A hierarchical multivariate denoising diffusion model.
\newblock \emph{Information Sciences}, 648:\penalty0 119623, 2023.

\bibitem[Wang et~al.(2023)Wang, Qi, Fang, and Sun]{wang2023hierarchical}
H.~Wang, L.~Qi, B.~Fang, and Y.~Sun.
\newblock Hierarchical visual policy learning for long-horizon robot manipulation in densely cluttered scenes.
\newblock \emph{arXiv preprint arXiv:2312.02697}, 2023.

\bibitem[Lugmayr et~al.(2022)Lugmayr, Danelljan, Romero, Yu, Timofte, and Van~Gool]{lugmayr2022repaint}
A.~Lugmayr, M.~Danelljan, A.~Romero, F.~Yu, R.~Timofte, and L.~Van~Gool.
\newblock Repaint: Inpainting using denoising diffusion probabilistic models.
\newblock In \emph{Proceedings of the IEEE/CVF conference on computer vision and pattern recognition}, pages 11461--11471, 2022.

\bibitem[Dhariwal and Nichol(2021)]{dhariwal2021diffusion}
P.~Dhariwal and A.~Nichol.
\newblock Diffusion models beat gans on image synthesis.
\newblock \emph{Advances in neural information processing systems}, 34:\penalty0 8780--8794, 2021.

\bibitem[Garrett et~al.(2020)Garrett, Lozano-P{\'e}rez, and Kaelbling]{garrett2020pddlstream}
C.~R. Garrett, T.~Lozano-P{\'e}rez, and L.~P. Kaelbling.
\newblock Pddlstream: Integrating symbolic planners and blackbox samplers via optimistic adaptive planning.
\newblock In \emph{Proceedings of the international conference on automated planning and scheduling}, volume~30, pages 440--448, 2020.

\bibitem[Huang et~al.(2023)Huang, Wang, Zhang, Li, Wu, and Fei-Fei]{huang2023voxposer}
W.~Huang, C.~Wang, R.~Zhang, Y.~Li, J.~Wu, and L.~Fei-Fei.
\newblock Voxposer: Composable 3d value maps for robotic manipulation with language models.
\newblock \emph{arXiv preprint arXiv:2307.05973}, 2023.

\bibitem[Minderer et~al.(2024)Minderer, Gritsenko, and Houlsby]{minderer2024scaling}
M.~Minderer, A.~Gritsenko, and N.~Houlsby.
\newblock Scaling open-vocabulary object detection.
\newblock \emph{Advances in Neural Information Processing Systems}, 36, 2024.

\bibitem[Kirillov et~al.(2023)Kirillov, Mintun, Ravi, Mao, Rolland, Gustafson, Xiao, Whitehead, Berg, Lo, et~al.]{kirillov2023segment}
A.~Kirillov, E.~Mintun, N.~Ravi, H.~Mao, C.~Rolland, L.~Gustafson, T.~Xiao, S.~Whitehead, A.~C. Berg, W.-Y. Lo, et~al.
\newblock Segment anything.
\newblock In \emph{Proceedings of the IEEE/CVF International Conference on Computer Vision}, 2023.

\bibitem[Sundermeyer et~al.(2021)Sundermeyer, Mousavian, Triebel, and Fox]{sundermeyer2021contact}
M.~Sundermeyer, A.~Mousavian, R.~Triebel, and D.~Fox.
\newblock Contact-graspnet: Efficient 6-dof grasp generation in cluttered scenes.
\newblock In \emph{2021 IEEE International Conference on Robotics and Automation (ICRA)}, 2021.

\end{thebibliography}

\clearpage
\appendix

\section{Related Works} \label{Sec: related works}

\subsection{Diffusion Models in Robotics}

Diffusion probabilistic models~\cite{ho2020denoising} are powerful generative frameworks that progressively denoise Gaussian distributions into target data distributions. These models have also been leveraged for imitation learning in robot policies and planners~\cite{zhao2024survey, urain2024deep}.

\textbf{Diffusion policies} utilize diffusion models to generate actions imitated from human demonstrations. Diffusion Policy~\cite{chi2023diffusion} employed vision-based diffusion models to predict actions for robot manipulation, significantly enhancing imitation learning performance in both simulation and real-world robots. Extending this, 3D diffusion policies~\cite{ze20243d, ke20243d} incorporated 3D point cloud observations to enrich spatial relationships and further improve manipulation performance. DNAct~\cite{yan2024dnact} applied 3D diffusion policies in multi-task environments, demonstrating the advantage of diffusion models in multi-modal distributions. \cite{hao2024language} utilized VLMs to generate keyframes to guide the diffusion policy. To address scalability in imitation learning, \cite{ha2023scaling} utilized visual-language models to generate diverse environments, paving the way for scalable robot foundation models.

\textbf{Diffusion planners} generate trajectories for mobile robots and manipulation tasks~\cite{carvalho2023motion}. Diffuser~\cite{janner2022planning} pioneered the use of diffusion models to synthesize goal-conditioned trajectories. Subsequent works enhanced diffusion planners with dynamic models~\cite{zhao2024diffusion}, potential fields~\cite{luo2024potential}, safety constraints~\cite{xiao2023safediffuser}, 3D observations~\cite{huang2023diffusion}, and multi-task settings~\cite{he2023diffusion}. Additionally, trajectory optimization via offline reinforcement learning was integrated into diffusion planners. Diffuser~\cite{janner2022planning} first employed classifier guidance for reward maximization, while \cite{yang2023planning} introduced inpainting methods for planning under uncertainty. Advances like discrete flow models~\cite{nisonoff2024unlocking} and classifier-free guidance~\cite{ajay2022conditional} further aligned trajectory prediction with reward optimization.

\textbf{Reinforcement learning (RL)} has also benefited from diffusion models, surpassing traditional MLP-based policies~\cite{zhu2023diffusion, zhang2024preferred}. Diffusion-QL~\cite{wang2022diffusion} used diffusion models as value networks, achieving superior value estimation. Diffusion-PPO~\cite{ren2024diffusion} employed diffusion models as policy networks optimized by PPO, significantly enhancing policy optimization. Similarly, Diffusion-AC~\cite{wang2024diffusion} revolutionized RL by integrating diffusion models into actor-critic frameworks. Further enhancements included Q-Score Matching~\cite{psenka2023learning} and out-of-distribution generalization~\cite{ada2024diffusion}. Offline RL integration, as seen in works like \cite{hansen2023idql} and \cite{zhang2024offline}, validated the ability of diffusion models to handle complex multi-modal distributions in policies and value networks.

\subsection{Hierarchical Trajectory Planners}

Hierarchical trajectory planning addresses the challenges of long-horizon trajectory generation~\cite{pateria2021hierarchical}. By decomposing trajectories into high-level (HL) subgoals and low-level (LL) segments, this approach reduces trajectory complexity, thereby improving planning efficiency and success rates. A critical challenge is maintaining coherent interactions between HL and LL levels.

%
\textbf{Long-horizon task planning} is particularly challenging in robot navigation and manipulation due to the need for both discrete grounded subtasks and fine-grained robot actions~\cite{chen2023llm, zhao2024survey}. Hierarchical planners are widely adopted in such scenarios~\cite{aceituno2022hierarchical}. Subgoal prediction~\cite{pertsch2020long} sequentially generates optimal subgoals and conditionally derives actions. High-level options~\cite{feng2024diffusion}, representing abstract skills, further expand the scope of long-horizon planning, enhancing generalizability and transferability. Recent developments in large language models (LLMs) and vision-language models (VLMs) have enabled task planning via general-purpose frameworks~\cite{yang2024guiding}. ReKep~\cite{huang2024rekep} leveraged VLMs to predict affordances and connections for long-horizon tasks. Hierarchical planners are also pivotal in tool-use planning, addressing interactions with functional tools and target objects~\cite{paulius2023long}.

\textbf{Hierarchical planning with diffusion models} has emerged as a promising approach. Subgoal Diffuser~\cite{huang2024subgoal} employed diffusion models for HL subgoal generation, guiding LL model predictive controllers. SkillDiffuser~\cite{liang2024skilldiffuser} used VQ-VAE for discrete HL skill generation, while LL diffusion models predicted manipulation videos. Skill-diffusion~\cite{kim2024robust} applied diffusion models for HL skill prediction in long-horizon tasks. Stitching Diffusion~\cite{kim2024stitching} utilized offline RL to combine sub-segments from diffusion models, and Option-based Diffusion~\cite{feng2024diffusion} planned HL abstracted options to guide LL diffusion planners.

Advanced hierarchical diffusion planners use diffusion models at both HL and LL levels. HDMI~\cite{li2023hierarchical} introduced a two-level diffusion planner, where HL subgoals were optimized by graph models, and LL segments were generated conditionally. SHD~\cite{chen2024simple} simplified this approach, using separate diffusion models for HL subgoals and LL segments. This baseline hierarchical diffuser (BHD) approach, however, suffers from a critical limitation: the independence of HL and LL levels prevents adjustments to erroneous subgoals. Motivated by this, we propose the Coupled Hierarchical Diffuser (CHD). Additionally, hierarchical diffusion planners have been applied in multi-agent games~\cite{wu2024diffusion}, multi-variable generation~\cite{zhang2023hierarchical}, and cluttered object environments~\cite{wang2023hierarchical}. Hierarchical structures have also been leveraged to accelerate sampling speeds~\cite{dong2024diffuserlite}.

\section{Baseline Hierarchical Diffusion Planner} \label{Appendix: BHD}
In Section~\ref{Sec: prelim}, we introduced the problem formulation of maximum-reward trajectory optimization, Diffuser structure and the baseline hierarchical diffuser (BHD). 
In the appendix, we present the complete formulation of BHD as a comparison to our proposed coupled hierarchical diffuser (CHD) method.

\subsection{Baseline Hierarchical Diffuser Structure} \label{Appendix: BHD-structure}
The Baseline Hierarchical Diffuser (BHD) is a straightforward application of diffusion models to hierarchical planning. HMDI~\cite{li2023hierarchical} and SHD~\cite{chen2024simple} are two instances of BHD.
BHD operates in two stages: a high-level (HL) planner generates sub-goals, and a low-level (LL) planner produces trajectory segments that meet these sub-goals via end-point inpainting~\cite{lugmayr2022repaint}.
\begin{equation} \label{Eqn: control as inference}
  \begin{aligned}
    p&(\btau^{g}, \btau^{x} \mid \mO = 1) \\
    & = p(\btau^{g}, \btau^{x}) p(\mO = 1 \mid \btau^{g}, \btau^{x}_{1:N}) \\
    & = p(\btau^{g}) p(\btau^{x} \mid \btau^{g}) p(\mO = 1 \mid \btau^{x}, \btau^{g}).
\end{aligned}  
\end{equation}
In Eqn.~\eqref{Eqn: control as inference}, BHD utilizes the Bayesian rule to decompose the control-as-inference~\cite{levine2018reinforcement} to optimize the trajectories to achieve `optimality'. The HL planner is a subgoal diffusion model $p(\btau^g)$ with classifier-guidance $p(\mO = 1 \mid \btau^{x}, \btau^{g})$, and the LL is a conditional diffusion model $p(\btau^{x} \mid \btau^{g})$. 

\textbf{High-Level (HL) Planner} generates a sequence of sub-goals $\btau^{g}$, and we impose that these sub-goals are ``optimal'', indicated by $\mathcal{O}=1$, as $p(\btau^{g}) \, p(\mathcal{O} = 1 \mid \btau^{g})$. Parameterizing these distributions, the HL reverse model is $p_{\theta^g}(\btau^g_{t-1} \mid \btau^g_{t}) \, p_{\theta}(\mO=1 \mid \btau^g_t)$, where $ p_{\theta^g}(\cdot) $ is the generative model for high-level sub-goals and $ p_{\theta}(\mathcal{O}=1 \mid \btau^g_t) $ models the probability that the sub-goal at time $t$ is optimal. Therefore, the HL planner is guided by the classifier and the reverse process is as~\cite{janner2022planning}:
\begin{equation} 
    \begin{gathered}
    p_{\theta^{g}}\left(\btau_{t-1}^{g} \mid \btau_{t}^{g}, \mathcal{O}_{1: N}\right) \approx \mathcal{N}\left(\btau_{t-1}^{g} ; \mu_{\theta^{g}}\left(\btau_{t}^{g}, t\right)+\Sigma^{t} \mathcal{J}\left(\mu_{\theta^{g}}\right), \Sigma^{t}\right) \\
    \mathcal{J}\left(\mu_{\theta^{g}}\right)= \nabla_{\mu_{\theta^g}} \left.\log p\left(\mathcal{O}_{1: N}=1 \mid \btau^{g}\right)\right|_{\btau^{g}=\mu_{\theta^g}}
    \end{gathered}
\end{equation}
\textbf{Low-Level (LL) Planner} generate trajectories $\btau^{x}$ that connect these sub-goals $\btau^{g}$ while satisfying endpoint constraints. We start with the conditional distribution $p(\btau^{x} \mid \btau^{g})$. 
The LL planner is defined by a diffusion process that refines initially noisy trajectories into coherent paths $p_{\theta^x}(\btau_{t-1}^x \mid \btau_t^x, \btau_0^g)$. 

\subsection{Loss function}

The training of HL and LL diffusion models are separated, and the variational bound of negative log-likelihood can be formulated as~\cite{li2023hierarchical}:
\begin{align}
    \mathbb{E}_q [-\log p_{\theta^g}(\btau^g_0) -\log p_{\theta^x}(\btau^x_0)] \leq \mathbb{E}_q \left[ -\log \frac{p_{\theta^g}(\btau^g_{0:T})}{q(\btau^g_{1:T} \mid \btau^g_0)} -\log \frac{p_{\theta^x}(\btau^g_{0:T} \mid \btau^g_0)}{q(\btau^x_{1:T} \mid \btau^x_0)}  \right] := L^{\text{BHD}}
\end{align}
We further expand the variational bound to derive the loss function as:
\begin{small}
\begin{equation}
\begin{aligned} \label{Eqn: BHD loss}
  L^{\text{BHD}}  & = \mathbb{E}_q \bigg[
\underbrace{D_{\text{KL}} \left( q(\btau_T^g \mid \btau_0^g) \| p(\btau_T^g) \right)}_{L_T^g} + \underbrace{D_{\text{KL}} \left( q(\btau_T^x \mid \btau_0^x) \| p(\btau_T^x) \right)}_{L_T^x} + \sum_{t=2}^T \underbrace{D_{\text{KL}} \left( q(\btau_{t-1}^g \mid \btau_t^g, \btau_0^g) \| p_\theta(\btau_{t-1}^g \mid \btau_t^g) \right)}_{L_{t-1}^g}  \\
& \qquad + \sum_{t=2}^T \underbrace{D_{\text{KL}} \left( q(\btau_{t-1}^x \mid \btau_t^x, \btau_0^x) \| p_\theta(\btau_{t-1}^x \mid \btau_t^x, \btau_0^g) \right)}_{L_{t-1}^x}  \underbrace{ - \log p_\theta(\btau_{0}^{g} \mid \btau_1^{g}) }_{L_0^g} \underbrace{ - \log p_\theta(\btau_{0}^{x} \mid \btau_1^{x}, \btau_1^{g}) }_{L_0^x} \bigg]
\end{aligned}
\end{equation}
\end{small}

In Eqn.~\eqref{Eqn: BHD loss}, the HL $\btau^g$ diffusion is independent, and the LL $\btau^x$ conditional diffusion is implemented via inpainting~\cite{lugmayr2022repaint}. Therefore, two diffusion models are independent in the training stage and could be trained separately.


\section{Joint Diffusion Model} \label{Appendix: JDM}

In section~\ref{Subsec: JDM}, we introduce the basic formulation of the joint diffusion model(JDM). JDM is a direct expansion of one diffusion model that jointly generates two variables as two coupled diffusion processes. We provide some details in this section.

\begin{figure}[ht]
    \centering
    \includegraphics[width=0.5\linewidth]{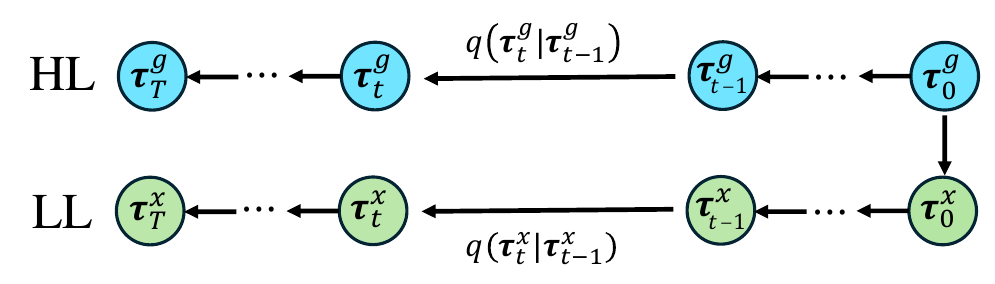}
    \caption{Probabilistic graph model of the forward process of joint diffusion model}
    \label{Fig: JDM forward}
\end{figure}

\textbf{Forward process}. Figure~\ref{Fig: JDM forward} shows the probabilistic graph model for the forward process of JDM. We consider two cases. 

When $t=0$, the LL trajectory is conditioned on the HL subgoals as:
$$q(\btau_t^g, \btau_0^x) = q(\btau_0^x | \btau_0^g) q(\btau^g_0);$$
when $t\geq0$, two diffusion processes have the Markov property that 
$$q(\btau_t^g | \btau_{t-1}^g, \btau_{t-1}^x) = q(\btau_t^g | \btau_{t-1}^g), \quad q(\btau_t^x | \btau_{t-1}^g, \btau_{t-1}^x) = q(\btau_t^x | \btau_{t-1}^x).$$
Therefore, the forward-nosing process of two joint variables is
\begin{equation}
\begin{aligned}
&q(\btau^g_{1:T}, \btau^x_{1:T} | \btau_0^g, \btau_0^x) = \prod_{t=1}^T q(\btau_t^g | \btau_{t-1}^g, \btau_{t-1}^x) q(\btau_t^x | \btau_{t-1}^g, \btau_{t-1}^x)  \\
& \qquad = \prod_{t=1}^T q(\btau_t^g | \btau_{t-1}^g) q(\btau_t^x | \btau_{t-1}^x) = q(\btau^g_{1:T} | \btau_0^g) q(\btau^x_{1:T} | \btau_0^x)
\end{aligned}
\end{equation}

$q(\btau^g_{1:T} | \btau_0^g)$ and $q(\btau^x_{1:T} | \btau_0^x)$ can be regard as two independent forward processes.

\textbf{Reverse process}. Figure~\ref{Fig: graph model}(b) shows the probabilistic graph model for reverse process of JDM. We can easily expand the joint distribution using Bayes Theorem as,
\begin{equation}
\begin{aligned}
    p(\btau^g_{0:T}, \btau^x_{0:T}) & =   p(\btau^g_T, \btau^x_T) \prod_{t=1}^T p_\theta(\btau_{t-1}^g, \btau_{t-}^x | \btau_{t}^g, \btau_{t}^x) \\
    & = p(\btau^g_T) p(\btau^x_T) \prod_{t=1}^T p_{\theta^g}(\btau_{t-1}^g | \btau_{t}^g, \btau_t^x) p_{\theta^x}(\btau_{t-1}^x | \btau_{t-1}^g, \btau^g_{t}, \btau_t^x) \\
    & = p(\btau^g_T) p(\btau^x_T) \prod_{t=1}^T p_{\theta^g}(\btau_{t-1}^g | \btau_{t}^g, \btau_{t}^x) p_{\theta^x}(\btau_t^x | \btau_{t-1}^g, \btau_t^x)
\end{aligned}
\end{equation}

In $p_{\theta^x}(\btau_t^x | \btau_{t-1}^g, \btau^g_{t} \btau_t^x)$, the $\btau^g_{t}$ is removed due to the Markov property.

\textbf{Loss function}. We derive the loss function with the variational bound of negative log-likelihood for the generative model as,
\begin{equation}
\mathbb{E}_q \left[ -\log p_\theta (\btau_0^g, \btau_0^x) \right] 
\leq \mathbb{E}_q \left[ -\log \frac{p_\theta (\btau_{0:T}^g, \btau_{0:T}^x)}{q(\btau_{1:T}^g, \btau_{1:T}^x | \btau_0^g, \btau_0^x)} \right] = L^{\text{JDM}}
\end{equation}

Then, we bring in the forward and reverse processes to derive~\cite{ho2020denoising}:
\begin{small}
\begin{equation} \label{Eqn: JDM training loss}
\begin{aligned}
L^{\text{JDM}} & = \mathbb{E}_q \left[ -\log \frac{p_\theta (\btau_{0:T}^g, \btau_{0:T}^x)}{q(\btau_{1:T}^g, \btau_{1:T}^x | \btau_0^g, \btau_0^x)} \right] \\
&= \mathbb{E}_q \bigg[ -\log p(\btau_T^g) - \log p(\btau_T^x) 
- \sum_{t=1}^{T} \log \frac{p_\theta(\btau_{t-1}^g | \btau_t^g, \btau_t^x) p_\theta(\btau_{t-1}^x | \btau_{t-1}^g, \btau_t^x)}
{q(\btau_t^g | \btau_{t-1}^g) q(\btau_t^x | \btau_{t-1}^x)} \bigg] \\
&= \mathbb{E}_q \bigg[ -\log p(\btau_T^g) - \log p(\btau_T^x) 
- \sum_{t=2}^{T} \log \frac{p_\theta(\btau_{t-1}^g | \btau_t^g, \btau_t^x) p_\theta(\btau_{t-1}^x | \btau_{t-1}^g, \btau_t^x)}
{q(\btau_t^g | \btau_{t-1}^g) q(\btau_t^x | \btau_{t-1}^x)} - \log \frac{p_\theta(\btau_0^g | \btau_1^g, \btau_1^x) p_\theta(\btau_0^x | \btau_0^g, \btau_1^x)}
{q(\btau_1^g | \btau_0^g) q(\btau_1^x | \btau_0^x)} \bigg] \\
&= \mathbb{E}_q \bigg[ \underbrace{D_{\text{KL}} \big( q(\btau_T^g | \btau_0^g) || p(\btau_T^g) \big)}_{L^g_T} 
+ \underbrace{D_{\text{KL}} \big( q(\btau_T^x | \btau_0^x) || p(\btau_T^x) \big)}_{L^x_T} + \sum_{t=2}^{T} \underbrace{D_{\text{KL}} \big( q(\btau_t^g | \btau_{t-1}^g) || p_\theta(\btau_{t-1}^g | \btau_t^g, \btau_t^x) \big)}_{L^g_{t-1}} \\ 
& \quad + \sum_{t=2}^T \underbrace{D_{\text{KL}} \big( q(\btau_t^x | \btau_{t-1}^x, \btau_0^x) || p_\theta(\btau_t^x | \btau_{t-1}^g, \btau_t^x) \big)}_{L^x_{t-1}} \underbrace{- \log p_\theta(\btau_0^g | \btau_1^g, \btau_1^x)}_{L^g_0} \underbrace{- \log p_\theta(\btau_0^x | \btau_0^g, \btau_1^x)}_{L^x_0} \bigg]
\end{aligned}
\end{equation}
\end{small}

In Eqn.~\eqref{Eqn: JDM training loss}, $L^g_T$ and $L^x_T$ are neglected since they do not have learnable parameters. $L^g_0$ and $L^x_0$ are the reconstruction terms that generate the original data distribution. In the experiments, we either use a Gaussian model for continuous variables or Bits Discretization~\cite{chen2022analog} for discrete variables. The most important losses $L^g_{t-1}$ and $L^x_{t-1}$ are denoising matching terms.


\section{Coupled Hierarchical Diffusion Planner} \label{Appendix: CHD}

In section~\ref{Subsec: CHD}, we propose the couple hierarchical diffusion algorithm (CHD) as an effective planner for three properties. The basic formulation of CHD is a simplification of JDM. The change is in the reverse process $p_{\theta^g}(\btau_{t-1}^g | \btau_{t}^g, \btau_t^x)$, $\btau_t^x$ is neglected so that the new reverse process is shown in Figure~\ref{Fig: graph model}(c). 
\begin{equation} \nonumber
p(\btau^g_{0:T}, \btau^x_{0:T}) = p(\btau^g_T) p(\btau^x_T) \prod_{t=1}^T p_{\theta^g}(\btau_{t-1}^g | \btau_{t}^g) p_{\theta^x}(\btau_{t-1}^x | \btau_{t-1}^g, \btau_t^x)
\end{equation}

\subsection{Loss function and parametrization} \label{Appendix: CHD loss}

We can derive the variational bound and loss function analogous to JDM (Eqn.~\ref{Eqn: JDM training loss}) as,
\begin{small}
\begin{equation} \label{Eqn: CHD training loss}
\begin{aligned}
L^{\text{CHD}} & = \mathbb{E}_q \bigg[ \underbrace{D_{\text{KL}} \big( q(\btau_T^g | \btau_0^g) || p(\btau_T^g) \big)}_{L^g_T} 
+ \underbrace{D_{\text{KL}} \big( q(\btau_T^x | \btau_0^x) || p(\btau_T^x) \big)}_{L^x_T} + \sum_{t=2}^{T} \underbrace{D_{\text{KL}} \big( q(\btau_t^g | \btau_{t-1}^g) || p_\theta(\btau_{t-1}^g | \btau_t^g) \big)}_{L^g_{t-1}} \\ 
& \quad + \sum_{t=2}^T \underbrace{D_{\text{KL}} \big( q(\btau_t^x | \btau_{t-1}^x, \btau_0^x) || p_\theta(\btau_t^x | \btau_{t-1}^g, \btau_t^x) \big)}_{L^x_{t-1}} \underbrace{- \log p_\theta(\btau_0^g | \btau_1^g, \btau_1^x)}_{L^g_0} \underbrace{- \log p_\theta(\btau_0^x | \btau_0^g, \btau_1^x)}_{L^x_0} \bigg]
\end{aligned}
\end{equation}
\end{small}

Therefore, $p_{\theta^{g}}\left(\btau_{t-1}^{g} \mid \btau_{t}^{g}\right)$ and $p_{\theta^{x}}\left(\btau_{t-1}^{x} \mid \btau_{t-1}^{g}, \btau_{t}^{x}\right)$ are the HL and LL reverse diffusion models. We parameterize them as in DDPM~\cite{ho2020denoising}.

\begin{align}
& p_{\theta^{g}}\left(\btau_{t-1}^{g} \mid \btau_{t}^{g}\right)=\mathcal{N}\left(\btau_{t-1}^{g} ; \mu_{\theta^{g}}\left(\btau_{t}^{g}, t\right), \Sigma_{\theta^{g}}\left(\btau_{t}^{g}, t\right)\right) \\
& p_{\theta^{x}}\left(\btau_{t-1}^{x} \mid \btau_{t-1}^{g}, \btau_{t}^{x}\right)=\mathcal{N}\left(\btau_{t-1}^{x} ; \mu_{\theta^{x}}\left(\btau_{t}^{x}, \btau_{t-1}^{g}, t\right), \Sigma_{\theta^{x}}\left(\btau_{t}^{x}, \btau_{t-1}^{g}, t\right)\right)
\end{align}

The exact loss function can be written as,
\begin{equation} \label{Equ: CHD parameterization loss}
\mathcal{L}_{\text{Diff}} = \mathbb{E}_{\btau_0^g, \btau_0^x, \epsilon^g, \epsilon^x, t} 
\left[ \left\| \epsilon^g - \epsilon^g_{\theta^g}(\btau_t^g, t) \right\| 
+ \left\| \epsilon^x - \epsilon^x_{\theta^x}(\btau_{t-1}^g, \btau_t^x, t) \right\| \right]
\end{equation}

Note that HL and LL can be trained with one loss function or trained separately with their own losses. While in either way, all variables $\btau_0^g, \btau_0^x, \epsilon^g, \epsilon^x$ must be sampled in the same demonstration. In practice, we can train HL and LL in two GPU devices for acceleration.

\subsection{Coupled hierarchical classifier guidance} \label{Appendix: CHD classifier}

The coupled hierarchical classifier guidance enables LL feedback to HL because the classifier considers and adjusts both levels. We copy Eqn.~\eqref{Eqn: chd classifier guidance} here and apply the classifier guidance for HL and LL diffusion models.
\begin{equation} \nonumber
\begin{aligned}
& p\left(\btau^{g}_{0:T}, \btau^{x}_{0:T} \mid \mathcal{O}_{1:N}=1\right)  \propto  p\left(\btau^{g}_{0:T}, \btau^{x}_{0:T}\right) p\left(\mathcal{O}_{1:N}=1 \mid \btau^{g}_{0:T}, \btau^{x}_{0:T}\right) \\
& \quad = p(\btau^g_T) p(\btau^x_T) \prod_{t=1}^T p_{\theta^g}(\btau_{t-1}^g | \btau_{t}^g) p_{\theta^x}(\btau_{t-1}^x | \btau_{t-1}^g, \btau_t^x)  p_{\phi}(\mathcal{O}_{1:N}=1 | \btau^g_{t-1}, \btau^x_{t-1})
\end{aligned}
\end{equation}
\begin{small}
\begin{align}
& p_{\theta^{g}}\left(\btau_{t-1}^{g} \mid \btau_{t}^{g}, \mathcal{O}_{1:N}=1\right)  \approx \mathcal{N}\left(\btau_{t-1}^{g} ; \mu_{\theta^{g}}\left(\btau_{t}^{g}, t\right) +\lambda^{g} \Sigma^{t}  \mathcal{J}\left(\mu_{\theta^{g}}, \mu_{\theta^{x}}\right), \Sigma^{t}\right) \label{Eqn: classifier guidance HL} \\
& p_{\theta^{x}}\left(\btau_{t-1}^{x} \mid \btau_{t-1}^{g}, \btau_{t}^{x},   \mathcal{O}_{1:N}=1\right) \approx \mathcal{N}\left(\btau_{t-1}^{x} ; \mu_{\theta^{x}}\left(\btau_{t-1}^{g}, \btau_{t}^{x}, t\right) +\lambda^{x} \Sigma^{t}  \mathcal{J}\left(\mu_{\theta^{g}}, \mu_{\theta^{x}}\right), \Sigma^{t}\right) \label{Eqn: classifier guidance LL}
\end{align}
\end{small}
\begin{equation}
\mathcal{J}\left(\mu_{\theta^{g}}, \mu_{\theta^{x}}\right) = \nabla_{\mu_{\theta^g} } \log p_{\phi}(\mathcal{O}_{1:N}=1 \mid \btau^g,\btau^x) \mid _{\begin{smallmatrix}
    \btau^g=\mu_{\theta^g} \\ \btau^x=\mu_{\theta^x}
\end{smallmatrix}}
\end{equation}
$\lambda^g$ and $\lambda^x$ are the scaling factors of classifier guidance. One classifier simultaneously guides two diffusion processes, where LL trajectories control the HL goals.

\subsection{Asynchronous parallel generation} \label{Appendix: parallel}

CHD inherits the conditional generation framework from JDM, so $\btau^x_t$ conditions on $\tau^g_t$, making the parallel generation at the same time $t$ impossible. Nevertheless, we can build an asynchronous structure by rewriting the reverse process as,
\begin{small}
\begin{equation} \nonumber
\begin{aligned}
& p\left(\btau^{g}_{0:T}, \btau^{x}_{0:T} \mid \mathcal{O}_{1:N}=1\right) \propto p(\btau^g_T) p(\btau^x_T) \underbrace{p_{\theta^g}(\btau_{T-1}^g | \btau_{T}^g)}_{\mathcal{P}^g_T} \\
& \cdot \prod_{t=1}^{T-1} \bigg[ \underbrace{p_{\theta^g}(\btau_{t-1}^g | \btau_{t}^g) p_{\theta^x}(\btau_{t}^x | \btau_{t}^g, \btau_{t+1}^x)  p_{\phi}(\mathcal{O}_{1:N}=1 | \btau^g_{t}, \btau^x_{t})}_{\mathcal{P}^{g,x}_{t-1}} \bigg] 
\underbrace{p_{\theta^x}(\btau_{0}^x | \btau_{0}^g, \btau_1^x)  p_{\phi}(\mathcal{O}_{1:N}=1 | \btau^g_{0}, \btau^x_{0})}_{\mathcal{P}^x_0}
\end{aligned}
\end{equation}    
\end{small}

Then, the new paired variables are $(\btau^g_{t-1}, \btau^x_t)$. More details are introduced in section~\ref{Subsec: CHD}, while one important change is the asynchronous classifier guidance on HL diffusion.

\begin{equation} \label{Eqn: appendix asy classifier guidance}
p_{\theta^g}(\btau_{t-1}^g | \btau_{t}^g) p_{\phi}(\mathcal{O}_{1:N}=1 | \btau^g_{t}, \btau^x_{t}) = p_{\theta^{g}}\left(\btau_{t-1}^{g} \mid \btau_{t}^{g}, \mathcal{O}_{1:N}=1\right)
\end{equation}

As in Eqn.~\eqref{Eqn: appendix asy classifier guidance}, the reverse model generates $\btau^g_{t-1}$, while the classifier can only guide $\btau^g_t$, which makes a mismatch of classifier guidance. To solve this problem, we apply the chain rule of gradient to derive,

\begin{equation}
p_{\theta^{g}}\left(\btau_{t-1}^{g} \mid \btau_{t}^{g}, \mathcal{O}_{1:N}=1\right)  \approx \mathcal{N}\left(\btau_{t-1}^{g} ; \mu_{\theta^{g}}\left(\btau_{t}^{g}, t\right) +\lambda^{g} \Sigma^{t} \mathcal{J}^{\text{Asy}}\left(\btau^g_t, \mu_{\theta^{x}}\right), \Sigma^{t}\right) \label{Eqn: classifier guidance Asy HL}
\end{equation}

\begin{equation}
\begin{aligned}
\mathcal{J}^{\text{Asy}}\left(\btau^g_t, \mu_{\theta^{x}}\right) & = 
\nabla_{\mu_{\theta^g}} \log p_{\phi}(\mathcal{O}_{1:N}=1 \mid \btau^g,\btau^x) \mid _{\begin{smallmatrix}
    \btau^g=\btau^g_{t} \\ \btau^x=\mu_{\theta^x} 
\end{smallmatrix}} \\
& = \nabla_{\btau^g} \log p_{\phi}(\mathcal{O}_{1:N}=1 \mid \btau^g,\btau^x) \mid _{\begin{smallmatrix}
    \btau^g=\btau^g_{t} \\ \btau^x=\mu_{\theta^x} 
\end{smallmatrix}} 
\frac{\partial \btau^g_t}{\partial \mu_{\theta^g}} \\
& = \sqrt{1-\beta_t} \nabla_{\btau^g} \log p_{\phi}(\mathcal{O}_{1:N}=1 \mid \btau^g,\btau^x) \mid _{\begin{smallmatrix}
    \btau^g=\btau^g_{t} \\ \btau^x=\mu_{\theta^x} 
\end{smallmatrix}} 
\end{aligned}
\end{equation}

The partial derivative $\frac{\partial \btau^g_t}{\partial \mu_{\theta^g}} = \sqrt{1-\beta_t}$ represents the derivative of the forward model $q(\btau^g_t \mid \btau^g_{t-1})$. Therefore, we can easily use the classifier with $\btau^g_t$ to guide the generated $\btau^g_{t-1}$.

\subsection{Segment-wise generation}

In hierarchical planning problems, the primary function is to reduce the trajectory horizon for lower data complexity. A common approach is dividing the trajectory into sub-segments and corresponding subgoals~\cite{chen2024simple}. In the CHD formulation, we apply the segment-wise generation for both LL planner (Eqn.~\eqref{Eqn: chd LL segment}) and hierarchical classifier (Eqn.~\eqref{Eqn: chd classifier segment}). In the following, we show the changed loss functions for segment-wise generation.

\begin{equation} \nonumber
p_{\theta^{x}}\left(\btau_{t-1,1: N}^{x} \mid \btau_{t-1}^{g}, \btau_{t, 1: N}^{x}\right) =\prod_{i=1}^{N} p_{\theta^{x}}\left(\btau_{t-1, i}^{x} \mid g_{t-1, i}, \btau_{t, i}^{x}\right)
\end{equation} 

\begin{equation}  \label{Eqn: CHD diff parameterization loss}
\begin{aligned}
\mathcal{L}_{\text{Diff}} & = \mathbb{E}_{\btau_0^g, \btau_0^x, \epsilon^g, \epsilon^x, t} 
\left[ \left\| \epsilon^g - \epsilon^g_{\theta^g}(\btau_t^g, t) \right\| 
+ \left\| \epsilon^x - \epsilon^x_{\theta^x}(\btau_{t-1}^g, \btau_t^x, t) \right\| \right] \\
& = \mathbb{E}_{\btau_0^g, \btau_0^x, \epsilon^g, \epsilon^x, t} 
\left[ \left\| \epsilon^g - \epsilon^g_{\theta^g}(\btau_t^g, t) \right\| 
+ \sum^N_{i=1} \left\| \epsilon^x_i - \epsilon^x_{\theta^x}(g_{t-1,i}, \btau_{t,i}^x, t) \right\|
\right]
\end{aligned}
\end{equation}

Eqn.~\eqref{Eqn: CHD diff parameterization loss} is the training loss function with parameterized diffusion models with DDPM~\cite{ho2020denoising}. Due to the segmentation, $\epsilon^x = \left\{ \epsilon^x_i \right\}^N_{i=1}$ is the segment-wise noises, and $\epsilon^x_{\theta^x}(g_{t-1,i}, \btau_{t,i}^x, t)$ is the corresponding models for each segment. 

Similarly, we use the segment-wise classifier to optimize the hierarchical model and training losses are as follows,

\begin{equation} \nonumber
    p\left(\mathcal{O}_{1:N}=1 \mid \btau^{g}_{t}, \btau^{x}_{t, 1:N}\right) =\prod_{i=1}^{N} p\left(\mathcal{O}_{i}=1 \mid g_{t, i}, \btau^{x}_{t, i}\right) = \prod_{i=1}^{N} \exp \left( \sum_{k=(i-1)h}^{ih-1} r(s_k, a_k) \right)
\end{equation}

\begin{equation} \label{Eqn: CHD classifier parameterization loss}
\mathcal{L}_{\text{Classifier}} = \mathbb{E}_{\btau_0^g, \btau_0^x, t} 
\left[ \sum^N_{i=1} \left\| p_{\phi}(g_{t-1,i}, \btau_{t,i}^x, t) - \sum_{k=(i-1)h}^{ih-1} r(s_k, a_k) \right\| \right]
\end{equation}

Eqn.~\eqref{Eqn: CHD classifier parameterization loss} denotes that the classifier predicts the summation of reward in each segment. In the trajectory planning problems, the rewards are calculated by the spent time steps to achieve the subgoal. For instance, in the maze navigation tasks, the agent receives ``+1'' for each step upon reaching the subgoal. 

\subsection{Algorithm} \label{Appendix: overall algorithm}

We summarize the CHD's training and sampling algorithms as in Algorithm~\ref{Alg: CHD training} and \ref{Alg: CHD sampling}. 

\begin{algorithm}[h]
\caption{CHD Training}
\label{Alg: CHD training}
\begin{algorithmic}[1]
   \REPEAT
   \STATE Sample ${\btau^g_0, \btau^x_0} \sim q(\btau^g_0, \btau^x_0)$
   \STATE Sample $t \sim \text{Uniform}(\{1,\dots,T\}) $
   \STATE Sample $\epsilon^g \sim \mathcal{N}(\mathbf{0}, \mathbf{I}^g)$, $\epsilon^x \sim \mathcal{N}(\mathbf{0}, \mathbf{I}^x)$ 
   \STATE Calculate $\btau^g_t$ and $\btau^x_t$ by forward model $q(\tau^{\diamond}_t| \tau^{\diamond}_0) = \mathcal{N}(\tau^{\diamond}_t; \sqrt{\bar{\alpha}_t}\tau^{\diamond}_0, (1-\bar{\alpha}_t)\bm{I}^{\diamond}), \diamond=\{g,x\}$
   \STATE Train diffusion models $\nabla_{\theta^g,\theta^x} \mathcal{L}_{\text{diff}}$ in Eqn.~\eqref{Eqn: CHD diff parameterization loss}
   \STATE Train classifier for rewards $\nabla_\phi \mathcal{L}_{\text{Classifier}}$ with Eqn.~\eqref{Eqn: CHD classifier parameterization loss}
   \UNTIL{converged}
\end{algorithmic}
\end{algorithm}

\begin{algorithm}[h]
   \caption{CHD Sampling}
   \label{Alg: CHD sampling}
\begin{algorithmic}[1]
    \STATE $p\left(\btau_{T}^{g}\right) \sim \mathcal{N}\left(\btau_{T}^{g} ; \mathbf{0}, \mathbf{I}^{x}\right)$, $p\left(\btau_{T}^{x}\right) \sim \mathcal{N}\left(\btau_{T}^{x} ; \mathbf{0}, \mathbf{I}^{g}\right)$
    \STATE Calculate $\btau_{T-1}^g \sim p_{\theta^g}(\btau_{T-1}^g | \btau_{T}^g)$
    \FOR{$t=T-1, \dots, 1$}
    \STATE Calculate reverse processes $\btau_{t-1}^g \sim p_{\theta^g}(\btau_{t-1}^g | \btau_{t}^g)$ and $\btau_{t}^x \sim p_{\theta^x}(\btau_{t}^x | \btau_{t}^g, \btau_{t+1}^x)$ in parallel
    \STATE Calculate $p_{\phi}(\mathcal{O}_{1:N}=1 | \btau^g_{t}, \btau^x_{t})$ and apply classifier guidance by Eqn.~\eqref{Eqn: classifier guidance Asy HL} and \eqref{Eqn: classifier guidance LL}
   \ENDFOR
   \STATE Calculate $p_{\theta^x}(\btau_{0}^x | \btau_{0}^g, \btau_1^x)$ and apply LL classifier guidance $p_{\phi}(\mathcal{O}_{1:N}=1 | \btau^g_{0}, \btau^x_{0})$
   \STATE \textbf{Return} $\btau^g_0$ and $\btau^x_0$
\end{algorithmic}
\end{algorithm}

\subsection{Can JDM adapt to a hierarchical planner?} \label{Appendix: JDM as planner}

The joint diffusion model (JDM) is equivalent to one diffusion model with two variables, so it naturally satisfies the property 1. We simplify JDM and utilize three structures to achieve all 3 properties. This raises a question: 
Can we directly adapt JDM to achieve properties 2 and 3 without any simplification? Can we add parallel generation and segment-wise generation to JDM? The general answer is \textbf{No}.

As we stated in the CHD derivation, at time $t$, $\btau^x_t$ is conditioned on $\btau^g_t$, which forbids synchronous parallel generation. Therefore, we change the reverse step sequence to be $(\btau^g_{t-1}, \btau^x_t)$. However, for JDM the $\btau^g_{t-1}$ also conditions on $\btau^x_t$, meaning the inter-relation cannot be simply broken via asynchronous generation. And this triggers the idea of simplifying JDM to break the chain.

In addition, applying segment-wise generation to JDM cannot efficiently reduce the trajectory horizon. Similarly to CHD, we can divide the trajectory into sub-segments and the corresponding subgoals. This segmentation can be applied to LL planner and classifier, but cannot work on HL planner. Since the HL segment-wise generation becomes $p_{\theta^g}(g_{t-1,i} \mid g_{t,i}, \btau^x_{t,i})$, which only focuses on generating one subgoal and neglecting the planning of coarse, long-horizon subgoals for the whole trajectory. Therefore, the HL planner cannot adopt segment-wise generation and conditions with the whole long-horizon LL trajectory. On the contrary, CHD breaks the correlation of HL planner and LL trajectory in the diffusion model and utilizes a segment-wise classifier instead. 

In short, JDM is not naturally suitable for hierarchical planning. CHD adopts a simplification of JDM and modules for the hierarchical planner. Nevertheless, it is likely to have other methods to build hierarchical planners that also achieve 3 properties.

\subsection{Advantages of CHD over BHD} \label{Appendix: advantage}

In this section, we claim two advantages of CHD over the BHD. They can theoretically support and validate the enhanced performances of CHD in the long-horizon planning tasks.

\textbf{1. CHD better approximates the JDM.}

Comparing the variational bounds in \eqref{Eqn: JDM training loss}, Eqn.~\eqref{Eqn: BHD loss} and  \eqref{Eqn: CHD training loss}. We can regard both BHD and CHD as a simplifications of the JDM. However, we can prove that CHD's is ``closer'' to JDM than BHD in terms of the KL divergence. We first denote the Kullback–Leibler(KL)-divergence of CHD and BHD to JDM.

\begin{equation}
    D_{\text{KL}}(L^{\text{JDM}} \| L^{\text{BHD}}) = \mathbb{E}_q \left[ \sum^T_{t=1} \log \frac{p_{\theta^g}(\btau^g_{t-1} | \btau^g_t, \btau^x_t)}{p_{\theta^g}(\btau^g_{t-1} | \btau^g_t)} + \log \frac{p_{\theta^x}(\btau^x_{t-1} | \btau^g_{t-1}, \btau^x_t)}{p_{\theta^x}(\btau^x_{t-1} | \btau^g_{0}, \btau^x_t)} \right]
\end{equation}

\begin{equation}
    D_{\text{KL}}(L^{\text{JDM}} \| L^{\text{CHD}}) = \mathbb{E}_q \left[ \sum^T_{t=1} \log \frac{p_{\theta^g}(\btau^g_{t-1} | \btau^g_t, \btau^x_t)}{p_{\theta^g}(\btau^g_{t-1} | \btau^g_t)} \right]
\end{equation}

\begin{equation}
    D_{\text{KL}}(L^{\text{JDM}} \| L^{\text{BHD}}) - D_{\text{KL}}(L^{\text{JDM}} \| L^{\text{CHD}}) = \mathbb{E}_q \left[ \sum^T_{t=1} \log \frac{p_{\theta^x}(\btau^x_{t-1} | \btau^g_{t-1}, \btau^x_t)}{p_{\theta^x}(\btau^x_{t-1} | \btau^g_{0}, \btau^x_t)} \right]
\end{equation}

According to the imitation learning from demonstration, we know that 
$$ p_{\theta^x}(\btau^x_{t-1} | \btau^g_{t-1}, \btau^x_t) \geq p_{\theta^x}(\btau^x_{t-1} | \btau^g_{0}, \btau^x_t),$$
since $p_{\theta^x}(\btau^x_{t-1} | \btau^g_{t-1}, \btau^x_t)$ is the correct diffusion process and $p_{\theta^x}(\btau^x_{t-1} | \btau^g_{0}, \btau^x_t)$ is only a special case. Finally, we can validate that $D_{\text{KL}}(L^{\text{JDM}} \| L^{\text{BHD}}) \geq  D_{\text{KL}}(L^{\text{JDM}} \| L^{\text{CHD}})$, the CHD's variational bound is closer to JDM than BHD.

\textbf{2. CHD achieves higher trajectory optimality than BHD.}

We claim that CHD achieves a higher probability of maximum reward optimality than BHD. We first compare the formulation of trajectory optimization: 
\begin{align}
&\text{BHD:} \quad  p(\btau^{x}, \btau^{g} \mid O=1) 
= p(\btau^{g}) p_{\theta^{x}}(\btau^{x} \mid \btau^{g}) p(O=1 \mid \btau^{g}) \\
& \text{CHD:} \quad p(\btau^{x}, \btau^{g} \mid O=1) 
= p(\btau^{g}) p(\btau^{x} \mid \btau^{g}) p(O=1 \mid \textcolor{blue}{\btau^{x}}, \btau^{g})
\end{align}

Therefore, the claim can be verified by proving CHD's classifier achieves a higher probability of optimization than BHD $p(O=1 \mid \btau^g) \leq p(O=1 \mid \btau^x, \btau^g)$.
We define arbitrary LL trajectories $\tilde{\btau}^x$, and the conditional optimality probability can be marginalized as:
\begin{equation}
p(O=1 \mid \btau^g) = \int p(O=1 \mid \tilde{\btau}^x, \btau^g)p(\tilde{\btau}^x \mid \btau^g) d\btau^x
\end{equation}

We also know that the optimality (maximum reward) is determined by the true $\btau^x$ sampled from the demonstration so that the optimality condition on arbitrary $\tilde{\btau}^x$ is always less than the true $\btau^x$.
\begin{equation}
p(O=1 \mid \tilde{\btau}^x, \btau^g) \leq p(O=1 \mid \btau^x, \btau^g)
\end{equation}

Therefore, we marginalize both sides:
\begin{equation}
\int p(O=1 \mid \tilde{\btau}^x, \btau^g)p(\tilde{\btau}^x \mid \btau^g) d\tilde{\btau}^x 
\leq \int p(O=1 \mid \btau^x, \btau^g)p(\tilde{\btau}^x \mid \btau^g) d\tilde{\btau}^x
\end{equation}
Also, we know the integration of a probability is smaller than 1 $\int p(\tilde{\btau}^x \mid \btau^g) d\tilde{\btau}^x \leq 1$.
Bring this to the inequality and we finally prove that CHD achieves higher optimality than BHD. $p(O=1 \mid \btau^g) \leq p(O=1 \mid \btau^x, \btau^g)$.
Equality holds when the optimality \(O\) is independent from HL subgoals \(\btau^x\).

\clearpage
\section{Experiment Details} \label{Appendix: exp}
\subsection{Maze Navigation Experiments} \label{Appendix: maze navigation}

\begin{figure}[h]
    \centering
    \includegraphics[width=0.7\linewidth]{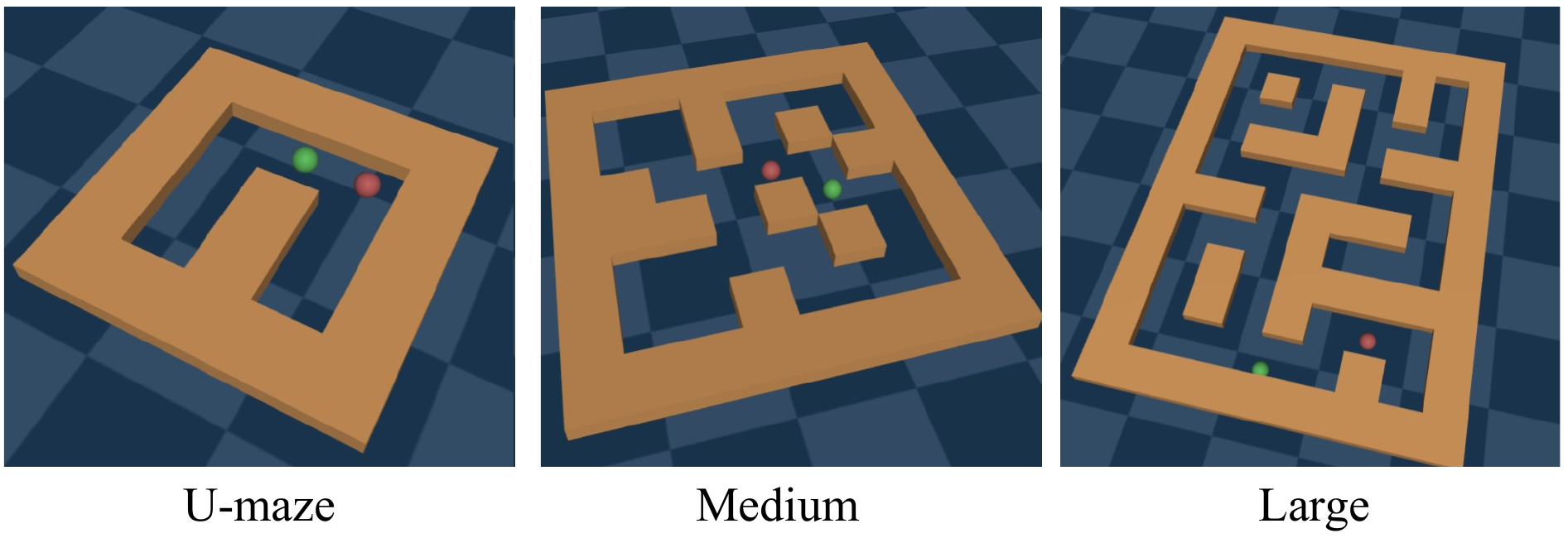}
    \caption{Maze navigation environment in D4RL. The agent start position (greed dot) is randomly chosen in the maze, and the goal position (red dot) is either fixed (Maze2D) or also randomized (Multi2D). The agent will get a ``$+1$'' reward when reaching the goal.}
    \label{Fig: maze nav illustration}
\end{figure}

\begin{figure}[h]
    \centering
    \includegraphics[width=0.7\linewidth]{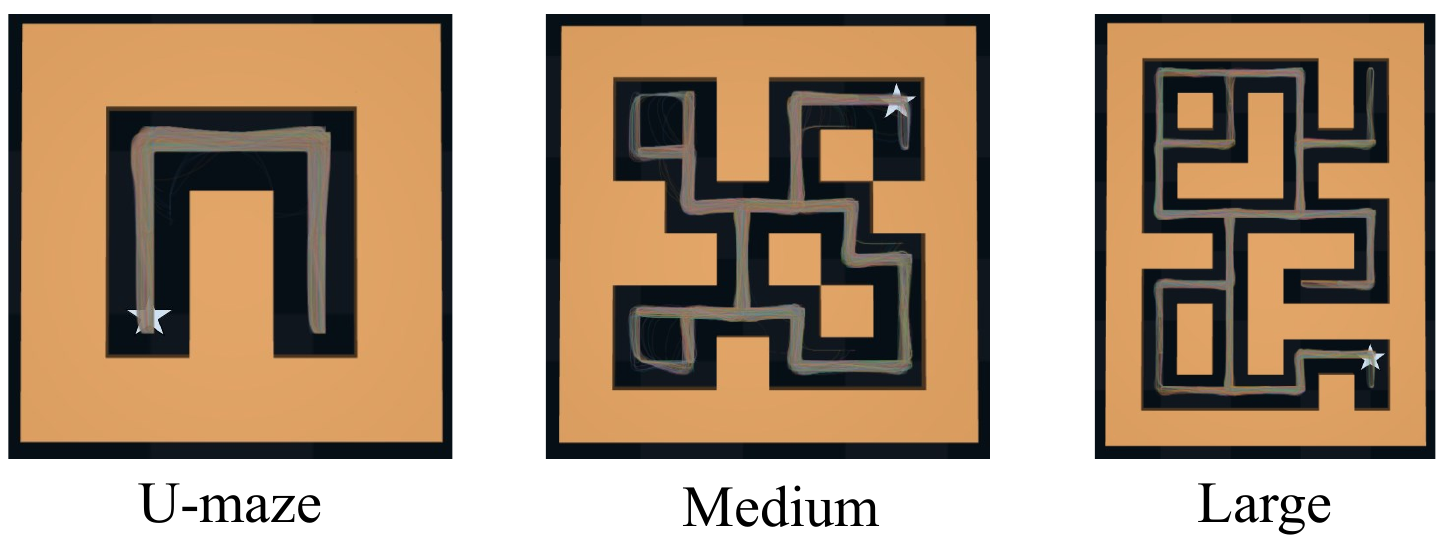}
    \caption{Demonstration in maze navigation environment. In the D4RL dataset, trajectories consist of states and actions. All trajectories do not have pre-defined start or goal position, therefore the trajectory segments may be redundant and sub-optimal. In addition, the agent can only observe the states rather than wall position.}
    \label{Fig: maze nav illu traj}
\end{figure}

\textbf{Environments and Datasets}

Maze navigation is a long-horizon trajectory planning environment in the D4RL dataset (Maze2D) ~\cite{fu2020d4rl}. The environment has three maze configurations with different sizes: U-maze, Medium and Large. In the simulator, the agent is initially placed in the maze and the target is to plan a feasible trajectory to reach the goal position. 
For each maze configuration, D4RL can either set a fixed goal position called ``Maze2D'' or a random goal position in the maze called ``Multi2D'', increasing the tasks' variance and requiring higher generalization ability (Fig.~\ref{Fig: maze nav illustration}).
In addition, the agent gets a ``$+1$'' reward when it reaches the vicinity of the goal position. So in a maze navigation environment, the planned trajectory should not only successfully reach the goal, but also reduce the time spent traveling to earn higher cumulated rewards.

In maze navigation, the agent is represented by a 2D point that moves in a closed maze. The state is $s = (x, y, v_x, v_y)$ and the action is $a = (v_x, v_y)$. D4RL provides offline demonstrations of the agent traveling in the maze. However, the trajectories do not have definite start or goal positions so they are sub-optimal for the goal-reaching problems (Fig.~\ref{Fig: maze nav illu traj}). In addition, since the demonstrations only provide state and action, the agent is unaware of the walls and obstacles in the maze, so the planners can only distill the feasible paths from the demonstrations rather than the wall position.

We finally evaluate the goal-reaching performance using the default method in D4RL. When the agent reaches the goal, it gets a ``$+1$'' reward and we calculate the overall reward in one episode. Then we applied maximin normalization and $\times 100$ to compare the results. Higher values mean that the trajectory used fewer steps to reach the goal. We run the experiments with 150 different seeds and report and mean and standard deviation of the results.

\clearpage
\textbf{Baselines}

We compare our CHD method with prior diffusion planner methods. 
\begin{itemize}[leftmargin=10pt]
    \item \textbf{Diffuser}~\cite{janner2022planning} is the initial diffusion model applied in trajectory planning. It leverages the diffusion model to predict both states and actions in the future. To enable goal-conditioned generation and maximum-reward trajectory optimization, Diffuser adopted endpoint inpainting~\cite{lugmayr2022repaint} method and classifier guidance~\cite{dhariwal2021diffusion}. 
    \item \textbf{Decision diffuser} (DD)~\cite{ajay2022conditional} improved Diffuser with diffusion transformer backbone and applied classifier-free guidance for trajectory optimization. 
    \item \textbf{BHD}~(Appendix~\ref{Appendix: BHD}) is the general baseline hierarchical diffuser method. It sequentially plans the HL subgoals and then uses subgoals as the endpoint condition to generate LL trajectory segments. The HL planner is guided by the maximum-reward classifier.
    \item \textbf{HDMI}~\cite{li2023hierarchical} is an improved instance of BHD. The difference is that HDMI pre-processes the dataset using graph search to generate better subgoals. Therefore, it performs better than BHD.
    \item \textbf{SHD} ~\cite{chen2024simple} is a simple implementation of BHD. It evenly segments the trajectories and uses endpoints are subgoals. With careful design, SHD further outperforms HDMI.
\end{itemize}

Since DD, SHD and HDMI did not provide the code for maze navigation, we directly consulted the results reported in their paper for comparison. 

\textbf{Implementation}

Our implementation was built on the code released in Diffuser \href{https://diffusion-planning.github.io/}{(code here)}. We first reproduced the Diffuser code and constructed the BHD and CHD algorithms. To ensure a fair comparison, we used U-net as the diffusion backbone and classifier guidance for reward maximization. The inpainting method was adopted for the endpoint-constraint of start and goal positions. 

\begin{figure}[h]
    \centering
    \includegraphics[width=0.8\linewidth]{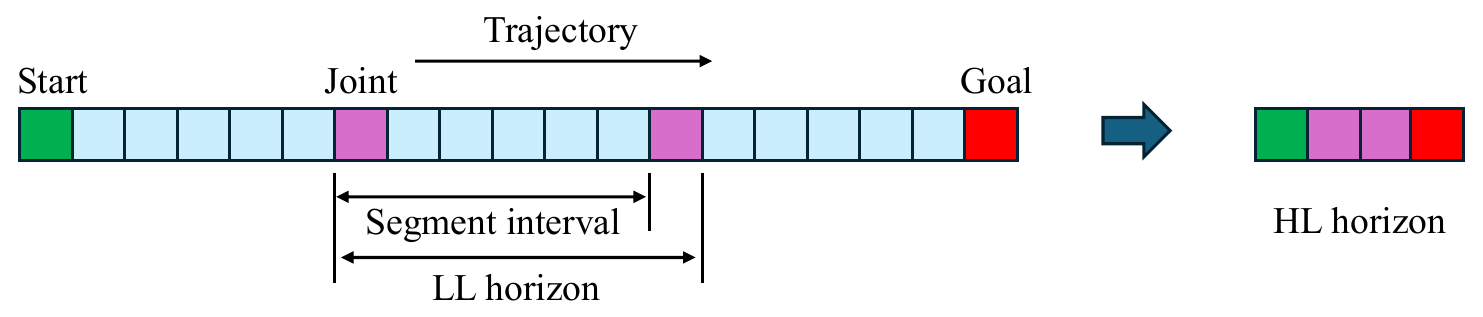}
    \caption{Trajectory segmentation for hierarchical diffuser training.}
    \label{Fig: maze seg illu}
\end{figure}

In the \textbf{training} stage (Fig.~\ref{Fig: maze seg illu}), we sampled fix-length trajectories from the offline demonstrations. Then we evenly divided the trajectories into sub-segments. The HL subgoals consist of the start, intermediate joint points, and the goal point, while LL trajectories are the segments. The segment-wise rewards are the ``negative number of steps before reaching the goal'' in the segment, which was identical to HDMI~\cite{li2023hierarchical}. This design can encourage the planner to ``speed up'' the agent and reduce traveling time. We applied this reward design in the HL classifier in BHD and the coupled hierarchical classifier in CHD.

In the \textbf{sampling} stage, we first initialize the agent and simulator to observe the start and goal positions. Then we applied the method to predict the further trajectory in a finite horizon. Note that the planning was open-loop and only executed at the initial state. In the downstream rollouts, a PD controller (implemented in Diffuser) was employed to track the planned trajectory. Finally, the simulator reported the overall rewards, namely the steps that the agent reaches the goal. Due to the different horizon lengths and maze configuration, D4RL normalized the overall reward with the distribution in the demonstrations. We followed the tradition and reported the experimental results in Table~\ref{Fig: maze results}.

\begin{table}[h]
    \centering
    \resizebox{.73\textwidth}{!}{
    \begin{tabular}{lcccccccc}
        \toprule
         & \multirow{2}{*}{\makecell[c]{Episode \\ length}} 
         & \multirow{2}{*}{\makecell[c]{Segment$^2$ \\ interval}} 
         & \multicolumn{3}{c}{Planner horizon} 
         & \multicolumn{3}{c}{Diffusion steps} \\ 
         \cmidrule(lr){4-6} \cmidrule(lr){7-9}
         & & & Single$^1$ & HL$^2$ & LL$^2$  & Single & HL & LL \\
        \midrule
        U-maze & 300 & 31 & 120  & 5 & 32  & 64  & 32 & 32 \\
        Medium & 600 & 31 & 320 & 12 & 32  & 256 & 32 & 32 \\
        Large  & 800 & 31 & 448 & 16 & 32  & 256 & 32 & 32 \\
        \bottomrule
    \end{tabular}
    }
    \vspace{2mm}
    \caption{Hyperparameters in the maze navigation experiments. $^1$ Single means the single-layer Diffuser. $^2$ In both BHD and CHD, the demonstration trajectories are segmented with the interval. HL subgoals are continuous joint endpoints; LL segments are the divided trajectories. }
    \label{tab:implementation_parameters}
\end{table}

We present the \textbf{hyperparameters} used in our implementation in Table~\ref{tab:implementation_parameters}. For Diffuser, we directly adopted the hyperparameters from the \href{https://diffusion-planning.github.io/}{official code}. Then, we adapted the hyperparameters in BHD and CHD to achieve a hierarchical structure. A notable modification was the reduction of diffusion steps, as the data distribution for shorter horizons is less complex and therefore easier to model. Additionally, the segment interval is defined as $\text{LL horizon} - 1$, since the endpoints correspond to two adjacent LL segments.

\begin{figure}[h!]
    \centering
    \includegraphics[width=0.97\linewidth]{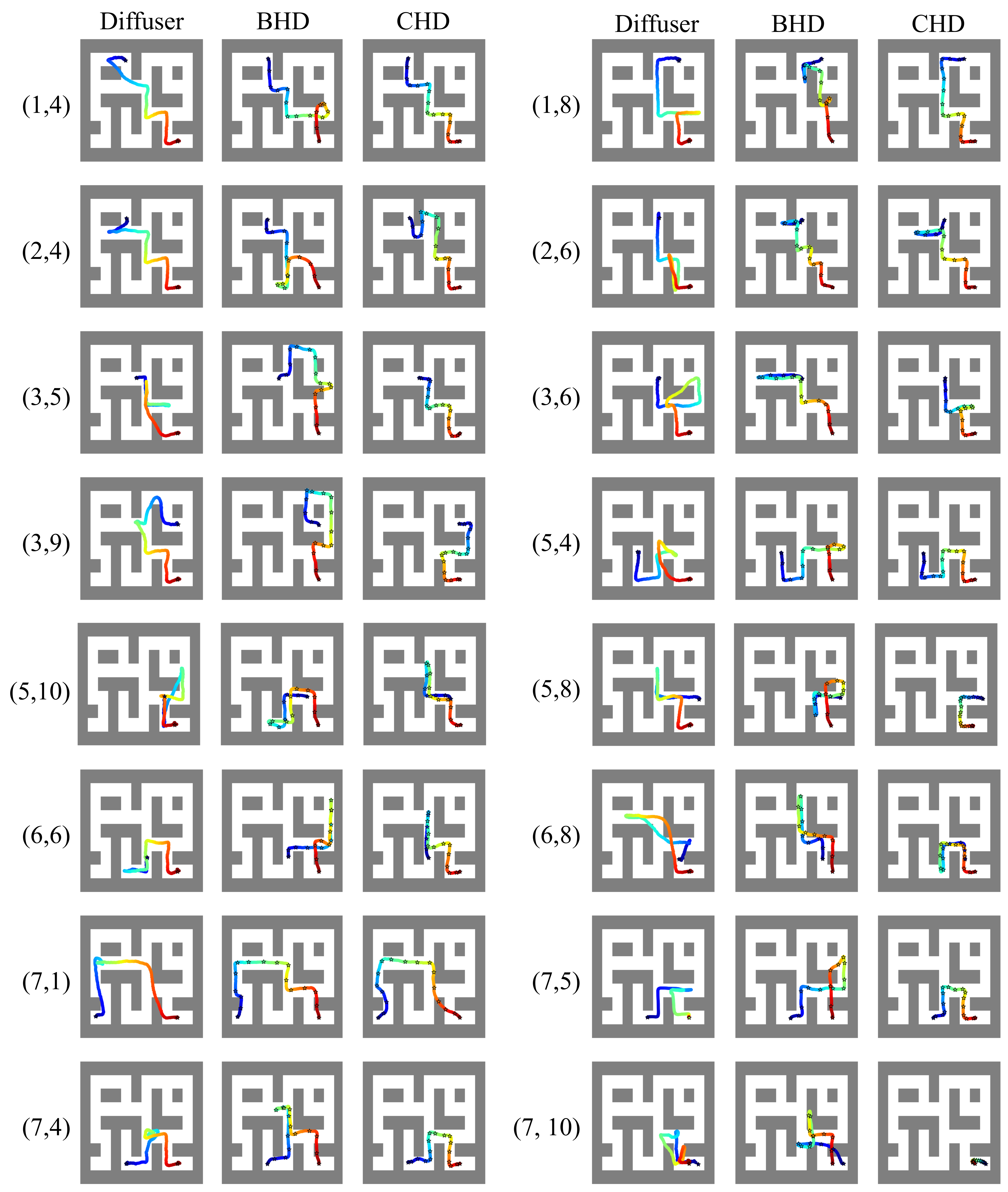}
    \caption{Visualization of maze navigation results in the Maze2D Large environment. The trajectory is from blue \textcolor{blue}{start} to red \textcolor{red}{goal}. The goal position is always on the bottom right, while the start position varies and is marked in each row.  $\bigstar$ represents intermediate sub-goals in BHD and CHD. }
    \label{Fig: maze more results}
\end{figure}

We \textbf{visualize} additional results in the Large Maze2D environment in Fig.~\ref{Fig: maze more results} (on the next page). In these visualizations, the goal position is fixed at the bottom-right corner, while the start position varies. Each row in the figure corresponds to a specific start position. The tasks can be categorized into three main scenarios:
\begin{enumerate}[leftmargin=15pt]
    \item \textbf{Start and goal are very far apart.} 
    The primary challenge in this scenario is to plan feasible, long-horizon, collision-free trajectories. Representative examples include $(1,4)$, $(1,8)$, $(2,4)$, $(2,6)$, $(3,5)$, and $(7,1)$. In general, the Diffuser can generate paths connecting the start and goal positions; however, it struggles with local control, resulting in collisions in cases like $(1,4)$ and $(3,5)$. BHD improves planning in cases like $(2,6)$ by incorporating long-horizon considerations in the HL. However, due to classifier guidance being applied only at the HL, it sometimes produces unachievable subgoals, as seen in $(1,8)$ and $(3,5)$. CHD demonstrates superior performance in most cases, producing near-optimal paths in $(1,4)$, $(1,8)$, and $(3,5)$, although it also exhibits undesirable results, such as in $(2,4)$. In the special case of $(7,1)$, where the planning horizon is insufficient to reach the goal, CHD shows an attempt to generate \textbf{smoother} paths through its optimization process.
    \item \textbf{Start and goal are near, with multi-modal paths.} 
    In these cases, the agent can initially move in multiple directions, with more than one valid path to the goal. Examples include $(3,6)$, $(3,9)$, and $(5,10)$. Here, the Diffuser performs poorly, generating unreasonable paths. BHD produces more reasonable paths but with noticeable local flaws. CHD significantly enhances performance, successfully generating feasible paths to the goal. However, some paths, such as $(5,10)$, still exhibit redundant turns.
    \item \textbf{Start and goal are much closer.} 
    Despite the shorter distances, these paths often involve numerous turns at corners, as seen in $(5,4)$, $(6,6)$, $(6,8)$, $(7,4)$, and $(7,10)$. In such cases, the planner must optimize trajectories to reduce travel steps while maintaining feasibility. Both the Diffuser and BHD generate generally feasible paths but suffer from redundant turns due to suboptimal demonstrations. The Diffuser's long-horizon trajectory modeling struggles with high variance, making trajectory optimization challenging. BHD, with its coarse HL goals, fails to provide feedback or adjust subgoals effectively. By contrast, CHD leverages coupled diffusion to align HL and LL, enabling superior path optimization that balances feasibility and near-optimality, as observed in $(7,4)$.
\end{enumerate}

From the experiments and comparisons, we conclude that CHD outperforms both the Diffuser and BHD methods in long-horizon maze navigation tasks. Across different planning scenarios, CHD consistently generates feasible, goal-reaching trajectories by imitating demonstrations and optimizing travel steps through coupled classifier guidance.

\clearpage
\subsection{Robot Task Planning Experiments} \label{Appendix: task planning exp}

\textbf{Environments and Datasets}

The robot task planning experiments were conducted using the Kitchen World simulator~\cite{yang2024guiding}, which is designed for learning-based algorithms in task planning~\cite{yang2022sequence, yang2024guiding} and supports integration with downstream solvers like PDDLStream~\cite{garrett2020pddlstream}. In Kitchen World, the PR2 dual-arm robot moves in the kitchen scenario, manipulating rigid and articulated objects to finish certain tasks.
In our experiments, we created a customized task-planning benchmark with extended horizons to evaluate performance. During the training phase, we utilized offline demonstrations to train the planners. In the sampling phase, the planner generated sub-goal states and actions, which were subsequently executed using the \href{https://github.com/caelan/pddlstream}{PDDLStream solver} as the policy. In short, the simplified PDDL maps the semantic action and targets to the robot and objects in the kitchen world simulation and plans trajectories to satisfy each primitive motion.

Task planning was formalized in a simplified PDDL space~\cite{garrett2021integrated}. The state space is represented as $s = [(\texttt{Object}, \texttt{Position})_n]$, where $n$ is the number of objects, and the action space is defined as $a = [(\texttt{Motion}, \texttt{Target})]$. The \texttt{Object} set includes all movable rigid and articulated objects in the environment, such as $[\texttt{Bowl}, \texttt{Pot}, \texttt{Cabinet}, \dots]$. The \texttt{Position} specifies the location or state of the object, e.g., $[\texttt{On-table}, \texttt{In-cabinet}, \texttt{Open}, \texttt{In-Bowl}, \dots]$. The \texttt{Motion} component of actions describes the robot's behavior at each step, encompassing $[\texttt{Pick}, \texttt{Place}, \texttt{Grasp}, \texttt{Push}, \texttt{Move}, \dots]$. The \texttt{Target} specifies the intended target of the motion, such as $[\texttt{Chicken}, \texttt{Cabinet}, \texttt{Right-Table}, \dots]$.

\begin{figure*}[h]
    \centering
    \includegraphics[width=0.97\linewidth]{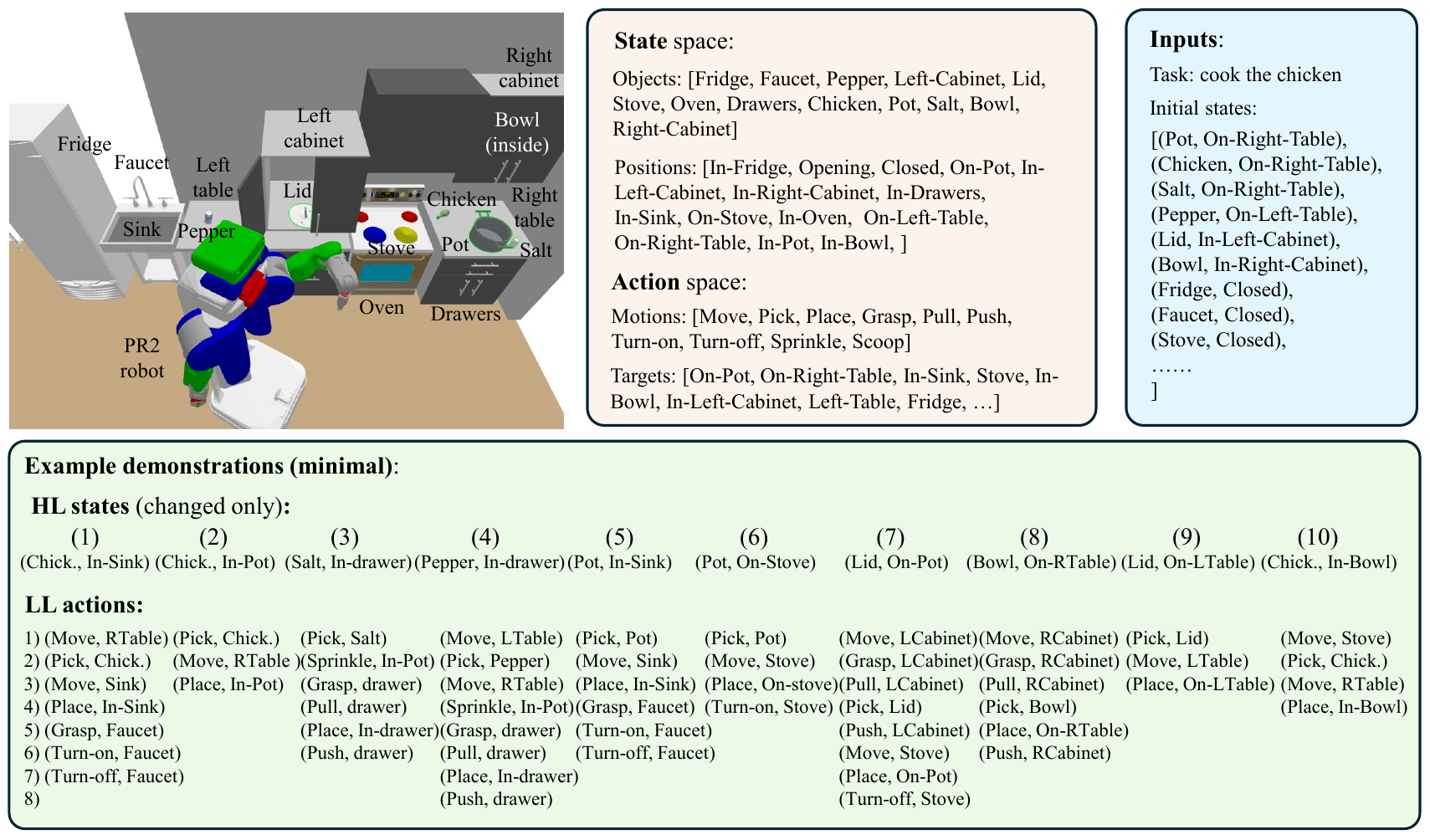}
    \caption{Illustration of task planning in the Kitchen World environment. Objects are randomly initialized in the scene, and a target task is specified. Based on the initial states, the program generates high-level (HL) subgoal states and low-level (LL) actions to complete the task. We use `Left' and `Right' as `L' and `R' for simplicity.}
    \label{Fig: kitchen illustration}
\end{figure*}

We constructed the dataset around the task of ``cooking meals'' within the kitchen environment, comprising 10 sub-tasks. Fig.~\ref{Fig: kitchen illustration} illustrates an example scene and demonstration. The state and action spaces, as well as initial states, are predefined in the scene. The program then generates HL subgoal states and LL actions required to accomplish the target task. The order of the 10 sub-tasks is interchangeable, provided they collectively achieve the target, which enhances the model's generalization capability in real-world scenarios. Each sub-task has a maximum of 10 actions and will receive a ``$+1$'' reward when the sub-task is finished. Therefore, the planner should generate successful subgoals and actions while reducing the number of actions for each sub-task.

To diversify demonstrations, we randomized the initial and terminal states of objects, varied the sub-task sequences, and introduced redundant behaviors. This approach ensured suboptimal demonstrations with extended trajectories. Based on the average trajectory lengths in the demonstrations, tasks were categorized into Easy (50 steps), Medium (70 steps), and Hard (90 steps). Additionally, a subset of tasks with fixed terminal states was defined as ``Single,'' while the full dataset with randomized terminal states was labeled ``Multi.'' 
The dataset consisted of $10,000$ training samples and $1,000$ validation samples. Finally, we evaluated the planned tasks in the simulator by measuring the number of completed tasks and the steps required to execute them.
We evaluate the planners' performance in terms of the number of successfully finished sub-tasks and the normalized steps to finish each sub-task. The preferred optimal planning should complete more sub-tasks and the cumulated action steps should be minimized.

\textbf{Baselines}

We applied baseline methods in the task planning environment. These baselines fall into three categories: auto-regressive transformers, single-layer diffusion models, and hierarchical diffusion models. The methods are:

\begin{enumerate}[leftmargin=15pt]
    \item \textbf{Large-Language Models (LLMs)}~\cite{yang2024guiding, huang2023voxposer}, such as ChatGPT-4o, are highly effective tools for task planning. They leverage general commonsense knowledge learned from web-scale datasets to break down complex tasks into sub-tasks. In our experiments, we adapted the prompts from \cite{yang2024guiding} and fed the training data to the LLM agent as in-context learning. This allowed the LLM agent to analyze tasks and generalize to new initial and goal states. During the sampling stage, we provided the initial and goal states, and ChatGPT-4o directly output all LL actions.
    \item \textbf{Transformer}~\cite{clinton2024planning} uses an auto-regressive structure to predict the next best token, making it naturally suitable for generating long-horizon trajectories. We employed the transformer as the backbone for task planning. Using the demonstrations, we trained the default ``gpt2'' transformer network in Huggingface~(\href{https://huggingface.co/docs/transformers/en/index}{Transformers}). During the sampling stage, the initial and target states were given as input, and a greedy policy was used to predict subsequent tokens until the end-of-sequence (EoS) token was reached.
    \item \textbf{Diffuser}~\cite{janner2022planning} was originally designed for continuous trajectory planning. To adapt it for planning in discrete token space, we applied the Bits Diffusion method~\cite{chen2022analog} to analogize the tokens. We then trained and validated the Diffuser for task planning, treating the initial and goal states as endpoint conditions.
    \item \textbf{BHD} (Baseline Hierarchical Diffusion, see Appendix~\ref{Appendix: BHD}) is the hierarchical diffusion baseline method. Similar to Diffuser, we applied Bits Diffusion to analogize tokens. We trained both HL and LL diffusers to generate HL subgoal states and LL actions. The HL planner used the initial and goal states as endpoint constraints, while the LL planner used the HL subgoals as endpoint constraints.
\end{enumerate}

Note that the LLMs, Transformer, and Diffuser baselines only generate LL actions, whereas both BHD and our proposed CHD plan generate HL subgoal states and LL actions.


\textbf{Implementation}

\begin{figure}[h]
    \centering
    \includegraphics[width=0.97\linewidth]{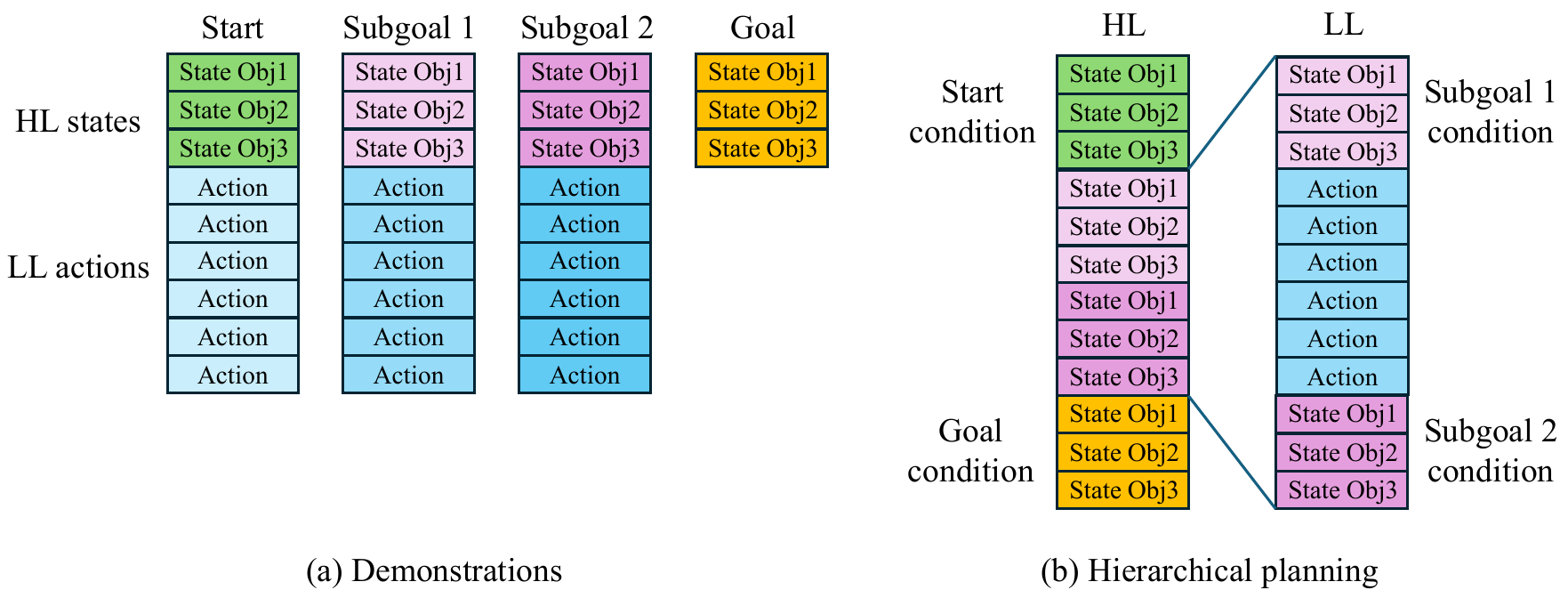}
    \caption{Illustration of demonstrations and hierarchical planning. (a) The demonstrations are divided into several subgoals with HL states and LL actions. (b) The HL and LL planners are trained with the reorganized data format. }
    \label{Fig: task planning training}
\end{figure}

We designed a paradigm for task planning with the hierarchical diffusion structure. In Fig.~\ref{Fig: task planning training}, the demonstrations are divided into sub-tasks with their HL subgoal states and LL actions. Then we re-organized the demonstrations to construct the HL and LL data. The HL planner has start and goal conditions, and it generates the intermediate subgoals. The LL planner conditions on two adjacent states and fills the actions that transit between two states. Finally, the actions were executed in the environment with the PDDLStream solver.

\begin{table}[h]
    \centering
    \resizebox{.85\textwidth}{!}{
    \begin{tabular}{lccccccccc}
        \toprule
         & \multirow{2}{*}{\makecell[c]{Average \\ demo. length}} 
         & \multirow{2}{*}{\makecell[c]{Episode \\ length}} 
         & \multirow{2}{*}{\makecell[c]{Segment \\ interval}} 
         & \multicolumn{3}{c}{Planner horizon} 
         & \multicolumn{3}{c}{Diffusion steps} \\ 
         \cmidrule(lr){5-7} \cmidrule(lr){8-10}
         & & & & Single$^1$ & HL$^2$ & LL$^2$  & Single & HL & LL \\
        \midrule
        Easy & 50 & 100 & 10 & 96  & 10 & 10   & 256  & 32 & 32 \\
        Medium & 70 & 120 & 10 & 112 & 10 & 12   & 256 & 32 & 32 \\
        Hard & 90 & 140 & 10 & 128 & 10 & 14   & 256 & 32 & 32 \\
        \bottomrule
    \end{tabular}
    }
    \vspace{2mm}
    \caption{Hyperparameters in the task planning experiments. $^1$ Single planners horizon means the single-layer methods like LLMs, Transformers and Diffuser.  $^2$ The HL planner generates 10 subgoal states; while LL planner generates the actions given adjacent states. }
    \vspace{2mm}
    \label{tab: task planning implementation_parameters}
\end{table}

Table~\ref{tab: task planning implementation_parameters} shows the hyperparameters used in the task planning experiments. In each environment, we set the episode length according to the average horizon of demonstrations. We set 10 HL subgoals for all tasks while the LL horizon is sufficiently longer than the demonstration. Therefore, the LL demonstrations are filled with padding tokens ``[PAD]'' at the end to satisfy the planning horizon. Comprising Diffuser and the hierarchical methods BHD and CHD, we also reduced the diffusion steps, since the hierarchical planners have much lower horizon and distribution complexity.

\begin{figure}[h]
    \centering
    \includegraphics[width=0.8\linewidth]{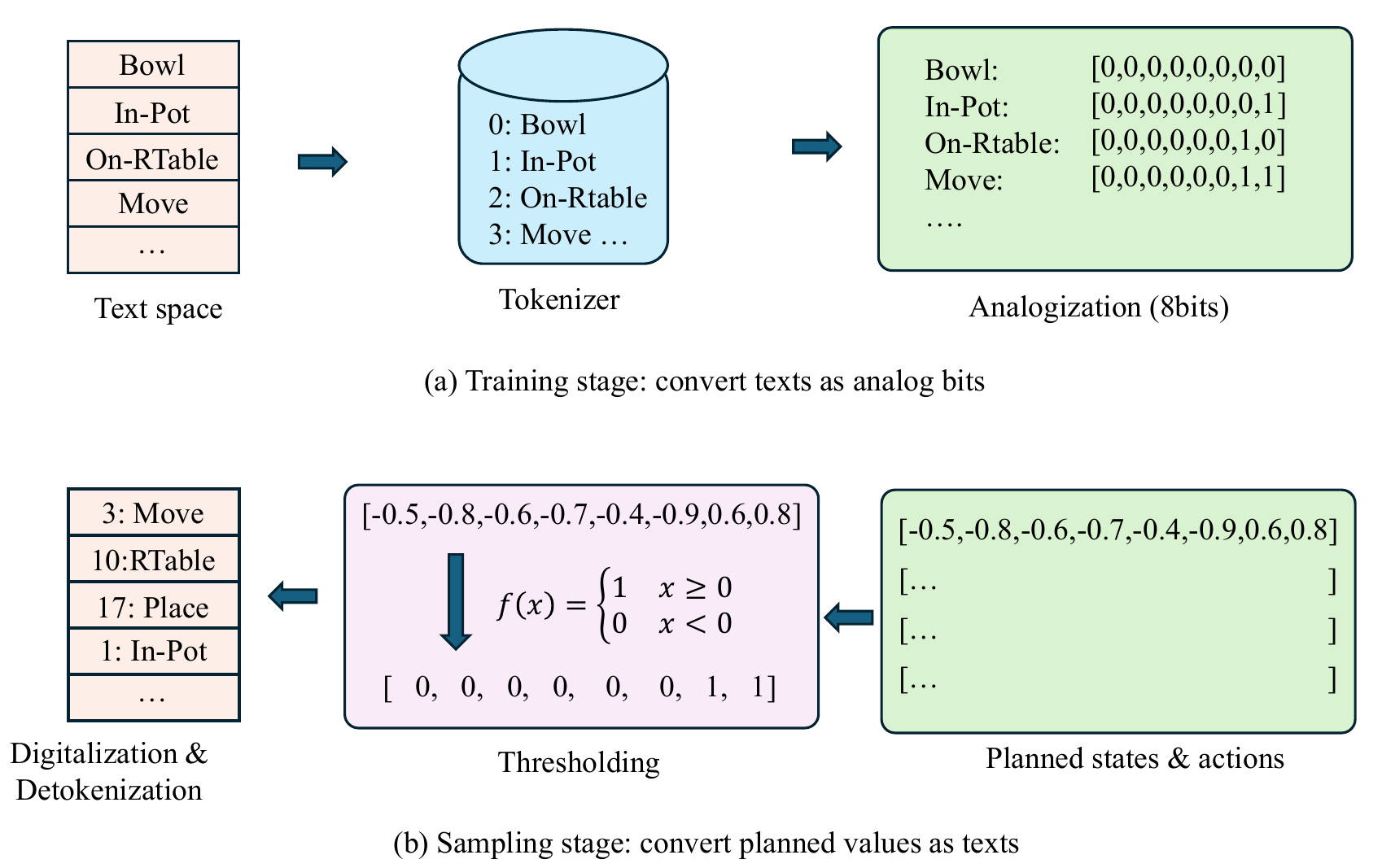}
    \caption{Illustration of Bits Diffusion for task planning experiments.}
    \label{Fig: bits illustration}
\end{figure}

We elaborate on implementing Bits Diffusion~\cite{chen2022analog} in the task planning experiments. In Fig.~\ref{Fig: bits illustration}, we first defined the whole state and action spaces, which were tokenized using the PreTrainedTokenizerFast in Huggingface (\href{https://huggingface.co/docs/transformers/en/main_classes/tokenizer}{code}). Then we converted the tokens as 8 Bits analog values. Since the state and action have two tokens, the diffusion model generates values with dimension ``horizon $\times 16$''. In the sampling stage, the planners sampled values in continuous space. And we converted them with thresholding to binary values. Finally, the bits were digitalized and de-tokenized back to the texts for the following steps.

\clearpage

\subsection{Real-robot Experiments} \label{Appendix: real robot}

Please see the \textcolor{red}{\textbf{video}} in the supplementary material or \href{https://sites.google.com/view/chd2025/home}{online website} for the real-robot experiment results.

\textbf{Environments and hardware}

\begin{figure}[h!]
    \centering
    \includegraphics[width=0.97\linewidth]{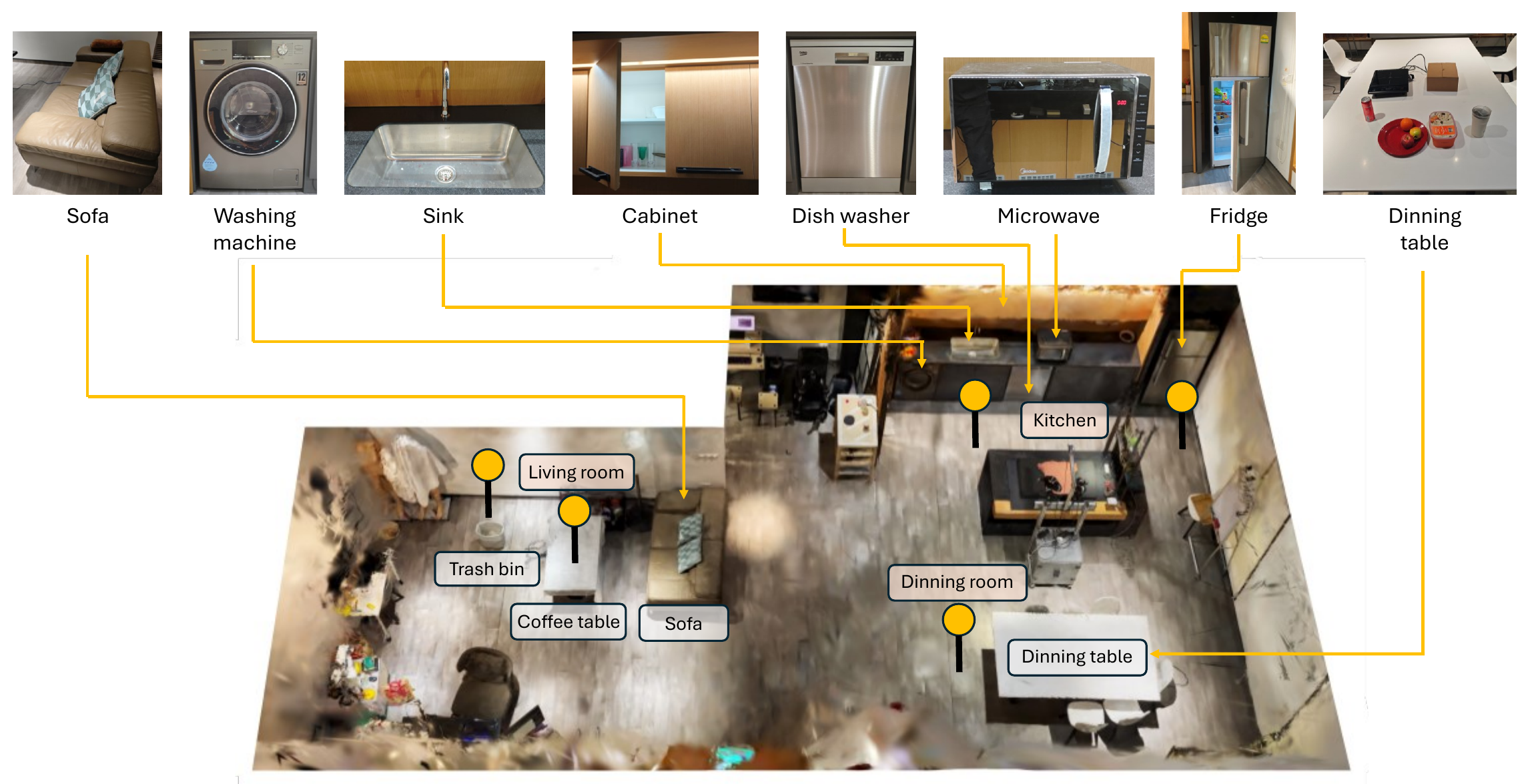}
    \caption{Top-down view of the real-robot experiment scenario.}
    \label{Fig: real topdown}
\end{figure}

We conduct our real-robot experiments using the Fetch mobile manipulator in a typical home environment comprising three rooms: the kitchen, living room, and dining room. Fig.~\ref{Fig: real topdown} presents a top-down view of the experimental setup.
%
The left side represents the \textbf{living room}, furnished with a sofa, coffee table, trash bin, chair, and cloth shelf. The \textbf{dining room}, located at the bottom right, contains a dining table surrounded by chairs. Items are initially placed on the dining table for subsequent interactions. The \textbf{kitchen}, in the top right, is the most critical area, equipped with various appliances and objects, including a washing machine, sink, dishwasher, microwave, fridge, and other tableware.

\begin{figure}[h!]
    \centering
    \includegraphics[width=0.65\linewidth]{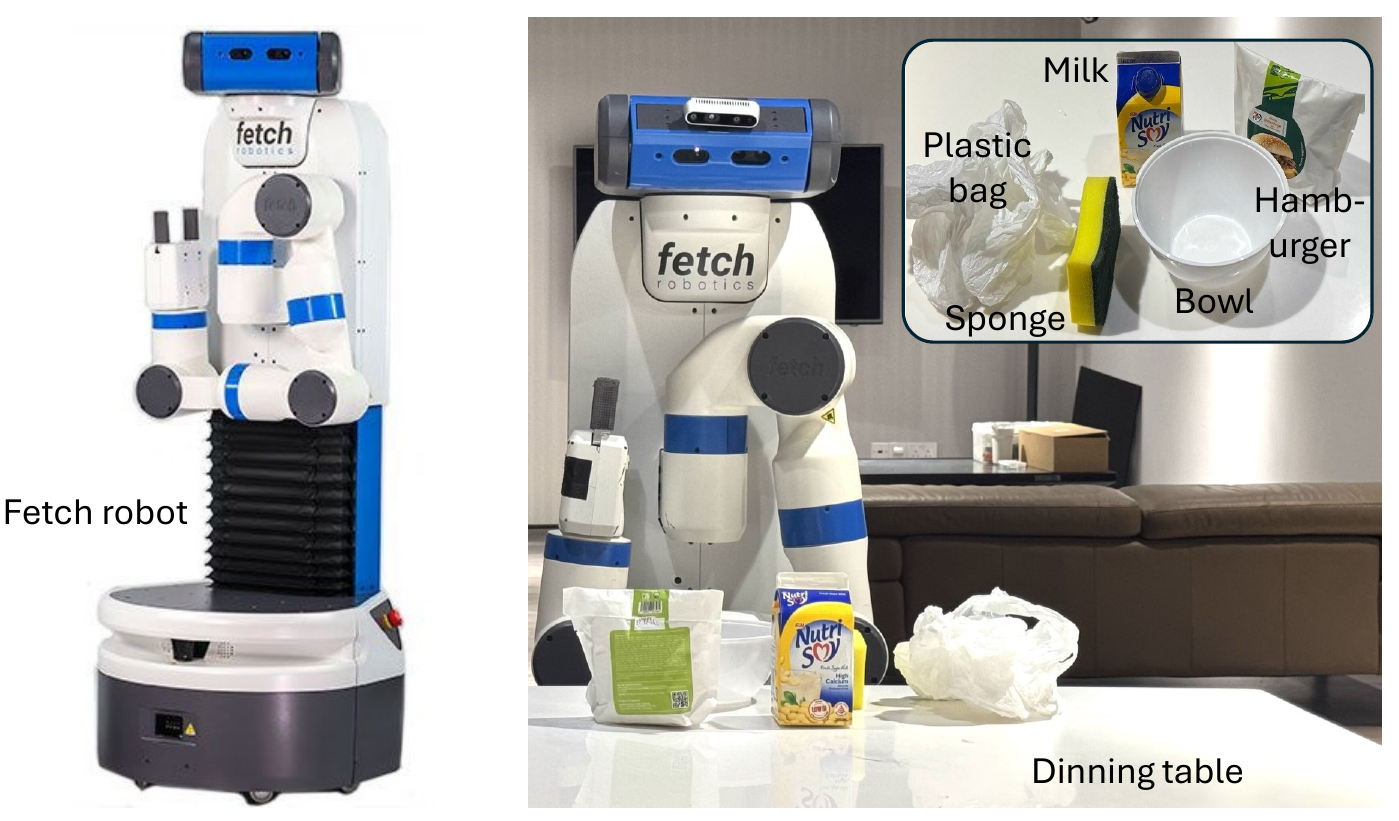}
    \caption{Fetch mobile robot and its environment. Objects are initially placed in clutter on the dining table, and the Fetch robot is tasked with organizing them.}
    \label{Fig: Fetch robot}
\end{figure}

The Fetch robot (Fig.~\ref{Fig: Fetch robot}) is a mobile manipulator designed for research and industrial automation. Developed by Fetch Robotics, it features a nonholonomic mobile base, a 7-degree-of-freedom (DoF) robotic arm, and a gripper, making it well-suited for tasks such as object manipulation, warehouse logistics, and human-robot interaction. In our experiments, The Fetch robot manipulates various rigid and articulated objects throughout the home.

\textbf{Task planner implementation}


We adopted the same task-planning method described in Appendix~\ref{Appendix: task planning exp}. 
First, we defined the state and action space of the mobile robot. 
We changed the actions with articulated objects as [\texttt{Open}, \texttt{Close}].
Next, we manually designed behaviors for randomly organizing items, such as storing objects in containers.
Based on these behaviors, we randomly generate a large task-planning dataset by forward sampling, ensuring comprehensive coverage of the possible state-action sequence distribution.
We used this dataset to train the CHD. Given any initial state and goal state, the CHD uses inpainting the HL subgoals and LL actions. We show the demonstration of the results in the video.
%

\begin{figure}[h!]
    \centering
    \includegraphics[width=0.97\linewidth]{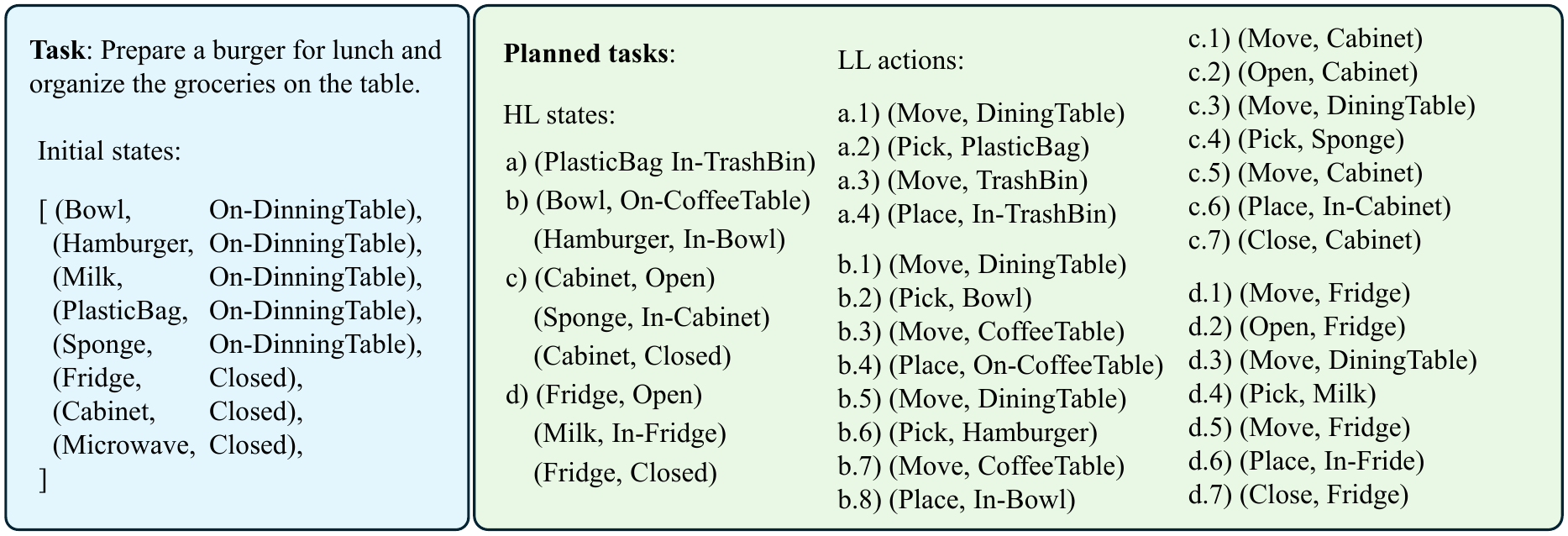}
    \caption{Example of real-robot task planning. According to initial states, CHD planned HL subgoal states and LL actions to complete the task.}
    \label{Fig: real task planning}
\end{figure}

Fig.~\ref{Fig: real task planning} shows an example of task planning. Given the initial states of objects, the CHD planner generated HL subgoals states and LL actions. We can categorize them into 4 sub-tasks. a) Drop the plastic bag; b) serve the hamburger; c) put sponge in the cabinet; d) put milk in the fridge. Finally, the robot adopted the real-robot controller to execute the actions to finish all tasks.

\textbf{Real-robot System Implementation}

All the computation for decision-making, perception, and low-level policy is done on a Linux workstation with an NVIDIA RTX 4090 GPU. The initial states are given at the beginning of the tasks for simplification, while the states could easily be constructed by using a VLM given the RGB observation.
The goal state, which defines where all objects should be, is generated by the LLM combined with the user's preference, e.g. user specifies the burger served in the bowl on the coffee table.
The robot is equipped with several navigation and manipulation skills: \texttt{pick, place, move, open, close}. These skills are implemented following the design of our previous mobile manipulation system \cite{xiao2024robi}.

The \texttt{pick} policy processes text queries in the format \texttt{pick(text)}. To identify and segment the target object, we leverage the pre-trained open-vocabulary detection model OWLv2 \cite{minderer2024scaling} in combination with the Segment Anything model \cite{kirillov2023segment}. The segmented object mask is then used in conjunction with the pre-trained grasping model Contact-GraspNet \cite{sundermeyer2021contact} to determine viable grasping poses. We refine these poses based on orientation constraints and select the one with the highest confidence score. A simple pre-grasp and grasp strategy is applied, with arm trajectories generated using MoveIT’s motion planning tools.

The \texttt{place} policy operates similarly to the \texttt{pick} and also supports text-based queries. Once the segmented point clouds of the target placement area are obtained, the placement position is computed: the center of the object is determined in the X-Y plane, while the height is set by adding 0.15 meters to the highest point in the segmented point clouds. For large, fixed objects or locations such as tables, counters, and trash bins, we simplify the process by using predefined placement locations. Additionally, for challenging placements (e.g., placing objects inside cabinets or fridges), we incorporate imitation learning for simplification.

For navigation, the \texttt{move} policy operates with predefined waypoints for all known locations. First, we generate an occupancy map using Gmapping and define navigation waypoints for known locations within the map. Path and motion planning are handled using the ROS Navigation Stack, which provides off-the-shelf algorithms for trajectory generation.

The \texttt{open} and \texttt{close} functions rely on imitation learning to execute complex actions, such as opening and closing a fridge or a cabinet. We collected an average of 50 human-teleoperated demonstrations per action using a VR controller on a real robot. These demonstrations were then used to train an Action Chunking with Transformers (ACT)~\cite{zhao2023learning} model. The model takes RGB-D images and the robot arm’s joint states as inputs to predict joint angle movement sequences, enabling smooth execution of these tasks.

\end{document}